\pdfoutput=1
\documentclass{article} %
\usepackage{conference,times}

\usepackage{amsmath,amsfonts,bm}

\def\eqref#1{equation~\ref{#1}}

\def\1{\bm{1}}

\DeclareMathAlphabet{\mathsfit}{\encodingdefault}{\sfdefault}{m}{sl}
\SetMathAlphabet{\mathsfit}{bold}{\encodingdefault}{\sfdefault}{bx}{n}

\usepackage{hyperref}
\usepackage{url}
\usepackage{booktabs}
\usepackage{graphicx}
\usepackage{tabularx}
\usepackage{amssymb}
\usepackage{multirow}
\usepackage{mathtools}
\newcommand{\std}[1]{\scriptsize{$\pm$#1}}

\usepackage{titlesec}

\title{Temporally Centered SIGReg Improves \\LeWorldModel Representations \\for Robot Policy Learning}

\author{
Chang Liu \quad
Fei Suo \quad
Yanzhou Jin \quad
Zeyu Ping
 \\
\textbf{
Yusuke Iwasawa \quad
Yutaka Matsuo \quad
Yaonan Zhu$^{\dagger}$
} \\[1mm]
Graduate School of Engineering,
The University of Tokyo, Tokyo, Japan \\[1mm]
$^{\dagger}$Corresponding author:
\texttt{yaonan.zhu@weblab.t.u-tokyo.ac.jp}
}

\conffinalcopy %
\begin{document}

\maketitle
\begin{abstract}
    Recent work on LeWorldModel (LeWM) has shown that the Sketched Isotropic Gaussian Regularizer (SIGReg) enables stable end-to-end world model learning from pixels by regularizing the latent representation toward an isotropic Gaussian.
    While effective for latent-space planning, the representations learned by Raw LeWM are poorly suited for downstream robot policy learning.
    In this paper, through Monte Carlo analysis, we show that the Raw LeWM objective biases variance allocation toward the temporally persistent component, thereby suppressing the variance of the temporally centered residual.
    Consistent with this analysis, trained Raw LeWM representations exhibit suppressed residual variation and reduced decodability of robot state and dynamics, particularly gripper dynamics, which are crucial for robotic manipulation.
    To address this issue, we apply SIGReg to \textbf{temporally centered residuals} rather than to the whole latent representation.
    This simple change decouples persistent and residual variance allocation while retaining an effective anti-collapse property.
    On the LIBERO benchmark, our method improves downstream policy success on the Goal suite by 1.66× and raises the average success rate across all suites from 63.6\% to 83.8\%.
    Without external pretraining, it also outperforms both Diffusion Policy trained from scratch and the pretrained OpenVLA baseline.
    These results associate the variance-allocation bias of Raw LeWM with the downstream policy gap, and show that decoupling persistent and residual
    variation yields representations better suited for downstream robot policy learning.
    \href{https://ryuuchou17.github.io/tclewm/}{\underline{Project Page}}.
\end{abstract}

\section{Introduction}
    \label{sec:introduction}

    World models \citep{ha2018world,hafner2019learning,hafner2019dream} learn predictive latent representations of environment
    dynamics, allowing agents to anticipate the consequences of their
    actions and enabling planning and policy learning.
    Many existing approaches \citep{hafner2019learning,hafner2019dream,hafner2023mastering} rely on observation reconstruction, reward
    prediction, or other task-specific supervision.
    Although effective, these objectives may retain unnecessary visual
    details and complicate learning from large reward-free multi-task
    datasets \citep{Deng2021DreamerProRM,balestriero2024learning,hutson2024policy}.
    
    Joint-embedding Predictive Architecture \citep{assran2023self,assran2025v} offers an
    appealing alternative.
    LeWorldModel (LeWM) \citep{maes2026leworldmodel} jointly trains an encoder and an
    action-conditioned latent predictor directly from pixels.
    To prevent representation collapse, it uses the Sketched Isotropic
    Gaussian Regularizer (SIGReg) \citep{balestriero2025lejepa}, which applies an Epps--Pulley (EP)
    normality criterion \citep{epps1983test} to random latent projections and encourages the
    latent distribution to follow an isotropic Gaussian.
    This enables end-to-end world model learning without reconstruction,
    pretrained encoders, momentum targets, or reward supervision.

    However, preventing representation collapse does not guarantee representations that are well suited for downstream robot policy learning.
    On the LIBERO-Long suite \citep{liu2023libero}, behavior-cloning policies trained on Raw LeWM representations achieve only $26.1\%$ average success under independent single-task training.
    Increasing the scale of world model training partially improves downstream policy performance, reaching $44.7\%$ under suite-wise 10-task joint training and $52.0\%$ when the world model is trained jointly on all 40 LIBERO tasks, but does not eliminate this performance limitation.
    These observations motivate us to examine how the Raw LeWM training objective shapes the information retained in its learned representations.

    \begin{figure*}[t]
        \centering
        \includegraphics[width=\textwidth]{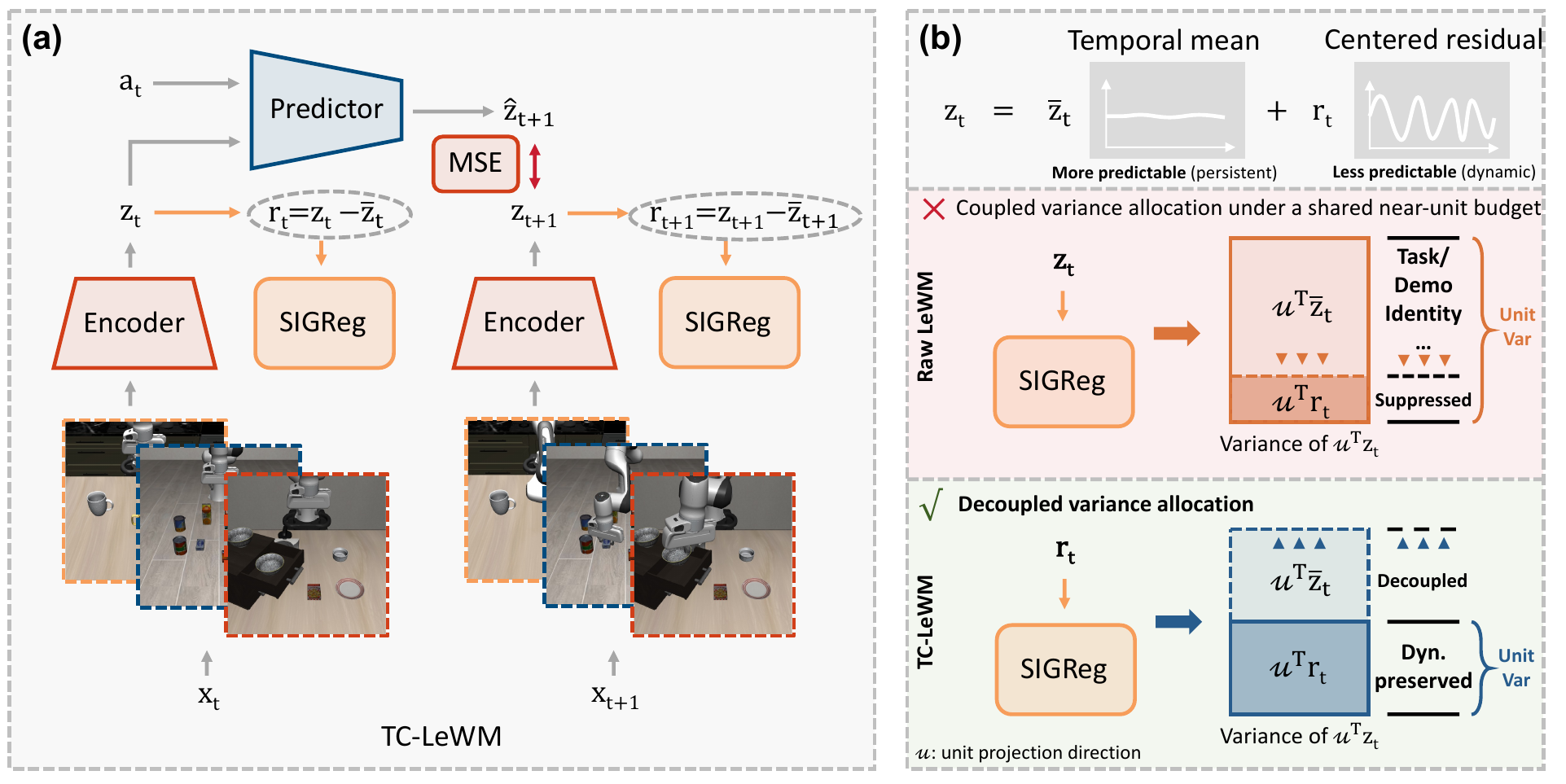}
        \caption{
            \textbf{Overview of TC-LeWM.}
            \textbf{(a)} TC-LeWM decomposes each latent representation as
            $\mathbf{z}_t=\bar{\mathbf{z}}_t+\mathbf{r}_t$, where
            $\bar{\mathbf{z}}_t$ is the temporally persistent component and
            $\mathbf{r}_t$ is the temporally centered residual. SIGReg is applied
            to $\mathbf{r}_t$ rather than directly to $\mathbf{z}_t$.
            \textbf{(b)} Along each regularized projection direction, the Raw LeWM objective
            couples the variance allocation of the persistent component and centered residual
            under a shared near-unit variance budget, whereas TC-LeWM decouples their variance
            allocation by regularizing the centered residual directly.
        }
        \label{fig:method_overview}
    \end{figure*}

    We trace this representation limitation to a variance-allocation bias in the objective of Raw LeWM.
    We first decompose the latent representation into a
    temporally persistent component and a temporally centered residual. Based on this
    decomposition, we develop a controlled Monte Carlo analysis of the Raw LeWM objective
    along individual SIGReg projection directions.
    The analysis reveals a mechanism. Along each projection direction, SIGReg
    encourages the projected latent variance toward the unit-variance target, while the 
    prediction objective favors allocating variance to more predictable temporally 
    persistent variation. Their interaction therefore couples the variance allocation 
    of the persistent component and centered residual under a shared near-unit variance 
    budget, biasing the allocation toward the persistent component and suppressing residual variance.

    To address this issue, we introduce \textbf{Temporally Centered LeWorldModel}
    (TC-LeWM), illustrated in Fig.~\ref{fig:method_overview}. TC-LeWM applies SIGReg
    to the temporally centered residual rather than directly to the original latent
    representation. This simple change decouples persistent and residual variance allocation
    while retaining an effective anti-collapse property.
    Across the four LIBERO suites, TC-LeWM raises the suite-wise 10-task average success rate of the downstream policy from $63.6\%$ to $83.8\%$. 
    When the world model is jointly trained on all 40 LIBERO tasks, TC-LeWM improves the average downstream policy success rate from $72.6\%$ to $85.4\%$.

    Consistent with our Monte Carlo analysis, Raw LeWM representations exhibit
    suppressed centered-residual variation, whereas TC-LeWM preserves substantially
    more variation in the residual component.
    Content attribution further shows that, under multi-task training, Raw LeWM allocates more variance to task and trajectory 
    identity, whereas TC-LeWM allocates relatively more variation to within-episode progress.
    Trajectory identity denotes the specific demonstration trajectory from which
    a frame is sampled and therefore captures trajectory-specific variation.
    The suppressed residual variation in Raw LeWM is also accompanied by lower
    decodability of robot state and dynamics, particularly gripper dynamics,
    which are crucial for closed-loop manipulation. 

    \noindent\textbf{Contributions.}
    \textbf{(i)} We introduce a temporal decomposition and controlled Monte Carlo analysis
    of the Raw LeWM objective, revealing its bias toward persistent variation and the
    resulting suppression of residual variance.
    \textbf{(ii)} We propose TC-LeWM, which applies SIGReg to the temporally centered
    residual to decouple persistent and residual variance allocation while retaining an
    effective anti-collapse property.
    \textbf{(iii)} Experimentally, TC-LeWM consistently improves the closed-loop performance
    of downstream behavior-cloning policies across three different training scales on LIBERO.
    We further analyze the learned representations through variance allocation, content attribution,
    and the decodability of robot state and dynamics.

    \section{Related Work}
    \label{sec:related_work}

    \noindent\textbf{Latent world models.}
    World models learn compact predictive representations of environment dynamics for planning and policy learning \citep{ha2018world,hafner2019dream,schrittwieser2020mastering,Hafner2020MasteringAW,hafner2023mastering}.
    Early latent-dynamics approaches showed that environment dynamics
    could be learned and used for control in compact latent spaces
    \citep{watter2015embed,hafner2019learning}.
    Subsequent stochastic state-space models combined deterministic memory
    with latent uncertainty, enabling control directly from pixel
    observations \citep{hafner2019dream,hafner2019learning}.
    Dreamer-style methods \citep{hafner2019dream,wu2022daydreamer, hafner2023mastering} further learned policies from trajectories
    imagined within the world model and extended this paradigm across
    continuous control, Atari, diverse domains, and physical robots.

    Most of these methods rely on observation reconstruction, reward
    prediction, or other task-specific objectives.
    Although effective, reconstruction objectives may retain
    control-irrelevant visual information, while reward-dependent
    objectives require supervision unavailable in reward-free datasets
    \citep{Fu2021LearningTI,Wan2023SeMAILED,Zhu2023RePoRM,balestriero2024learning,hutson2024policy,Wang2024AD3IA,
    Sobal2025LearningFR}.

    \noindent\textbf{Joint-embedding predictive learning.}
    Joint-Embedding Predictive Architectures (JEPAs) learn representations
    by predicting target embeddings directly in latent space rather than
    reconstructing raw observations
    \citep{assran2023self,dawid2024introduction}.
    I-JEPA \citep{assran2023self} instantiated this paradigm for images, and V-JEPA \citep{Bardes2024RevisitingFP,assran2025v} extended it to
    spatiotemporal representation learning from video.
    Related approaches have adapted latent prediction to
    action-conditioned world modeling:
    DINO-WM \citep{Zhou2024DINOWMWM} learns dynamics over frozen pretrained visual features for
    zero-shot planning, whereas PLDM \citep{Sobal2025LearningFR} jointly learns reconstruction-free
    latent dynamics from reward-free trajectories.
    However, non-contrastive joint-embedding objectives admit collapsed
    solutions, motivating mechanisms such as momentum target encoders,
    stop-gradient asymmetry, explicit variance regularization, and
    redundancy-reduction objectives
    \citep{Grill2020BootstrapYO,Chen2020ExploringSS,
    Bardes2021VICRegVR,Zbontar2021BarlowTS}.

    \noindent\textbf{LeWorldModel and anti-collapse regularization.}
    LeWM \citep{maes2026leworldmodel} enables end-to-end reconstruction-free world model
    learning directly from pixels. It jointly trains an encoder and an action-conditioned
    predictor using latent next-state prediction together with SIGReg
    \citep{balestriero2025lejepa}, which prevents representation collapse without pretrained
    encoders, momentum targets, observation reconstruction, or reward supervision.

    SIGReg is applied directly to latent representations and encourages their distribution
    toward an isotropic Gaussian through random one-dimensional projections. While effective
    for anti-collapse, how the Raw LeWM objective allocates variance across temporal
    components has received little attention. We show that the interaction between latent
    prediction and SIGReg couples the variance allocation of the temporally persistent
    component and centered residual, biasing the allocation toward persistent variation
    and suppressing residual variance. TC-LeWM addresses this issue by applying SIGReg
    to the temporally centered residual, thereby decoupling persistent and residual
    variance allocation while retaining an effective anti-collapse property.

    \begin{figure*}[t]
        \centering
        \includegraphics[width=\textwidth]{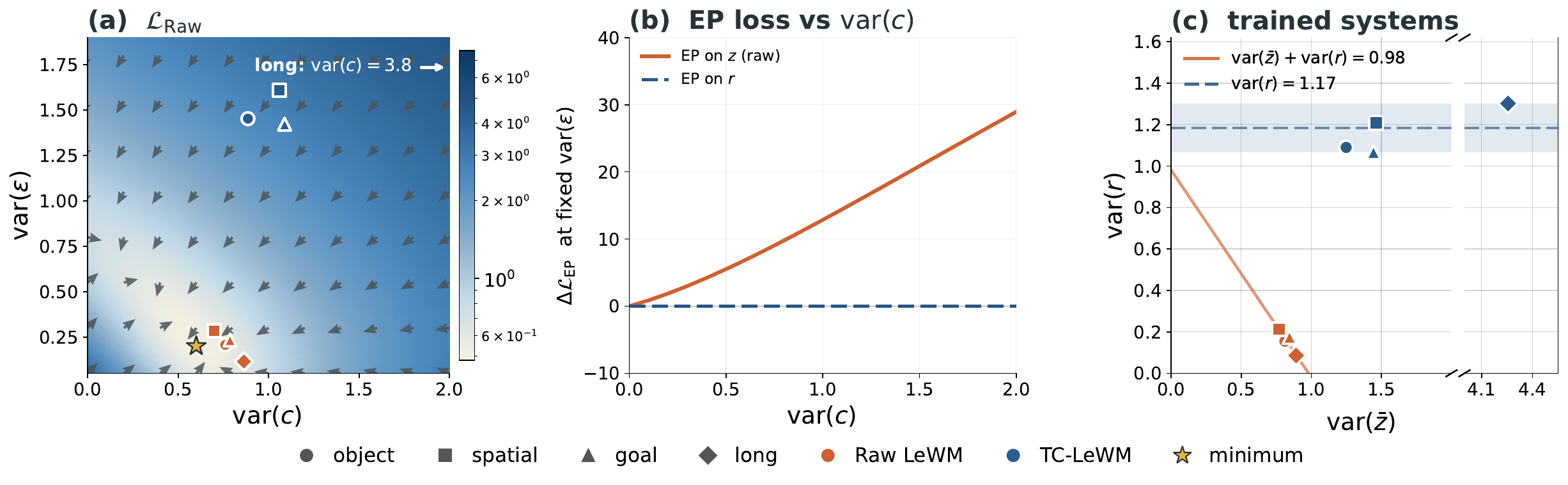}
        \caption{
            \textbf{Monte Carlo and empirical analysis of variance allocation in LeWM.}
            \textbf{(a)} The Raw LeWM objective $\mathcal{L}_{\mathrm{Raw}}$
            over the $(\mathrm{var}(c),\mathrm{var}(\varepsilon))$ plane, revealing
            a minimum biased toward persistent variation.
            \textbf{(b)} Change in EP loss relative to $\mathrm{var}(c)=0$ at fixed
            $\mathrm{var}(\varepsilon)$. Increasing persistent variance raises the
            EP loss on $x=c+\varepsilon$.
            \textbf{(c)} Empirical variance allocation of trained LIBERO representations.
            Raw LeWM exhibits a nearly fixed total variance dominated by the persistent
            component, whereas TC-LeWM maintains a stable residual variance with
            persistent variance free to vary.
        }
        \label{fig:mechanism}
    \end{figure*}

    \section{Why the Raw LeWM Objective Can Suppress Residual Variation}
    \label{sec:analysis}

    In this section, we first characterize the performance limitation of Raw LeWM
    across different training scales. We then introduce a temporal decomposition of
    the latent representation and, based on this decomposition, construct a projected
    mixture model for analyzing the Raw LeWM objective. Through a Monte Carlo
    analysis, we show how the interaction between latent prediction and SIGReg biases
    variance allocation toward temporally persistent variation and suppresses residual
    variance. 

    \subsection{Downstream Policy Performance Across Training Scales}
    \label{sec:task_scale}

    \noindent\textbf{World model training and downstream evaluation.}
    Consider a task set $\mathcal{T}=\{1,\ldots,N\}$, where task $i$ is
    associated with a demonstration dataset $\mathcal{D}_i$.
    Raw LeWM jointly trains an encoder and an action-conditioned latent
    predictor on the aggregated dataset
    $\mathcal{D}=\bigcup_{i=1}^{N}\mathcal{D}_i$ using
    \begin{equation}
        \mathcal{L}_{\mathrm{Raw}}
        =
        \mathcal{L}_{\mathrm{pred}}
        +
        \lambda\,\mathcal{L}_{\mathrm{SIGReg}}(\mathcal{Z}),
        \qquad
        \mathcal{Z}
        =
        \{\mathbf{z}_t
        \mid
        \mathbf{z}_t=f_\theta(\mathbf{x}_t)\}.
        \label{eq:lewm_objective}
    \end{equation}
    After world model training, the encoder is frozen and used for downstream
    robot policy learning 
    with a task-conditioned behavior-cloning policy based on flow matching~\citep{lipman2023flow}.

    \noindent\textbf{Limited benefit from training scale.}
    We evaluate Raw LeWM at three training scales of LIBERO-Long. When each of the ten tasks is trained
    independently, Raw LeWM achieves only $26.1\%$ average success. Joint training
    over the ten tasks raises the success rate to $44.7\%$, while
    training the world model jointly on all 40 LIBERO tasks further raises it to
    $52.0\%$. Thus, increasing the amount and diversity of training data substantially
    improves Raw LeWM, but does not eliminate its performance limitation.
    These results show that avoiding representation collapse alone does not guarantee
    representations suitable for downstream manipulation policy learning. 

    \subsection{Temporal Decomposition and Monte Carlo Analysis}
    \label{sec:mc_analysis}

    \noindent\textbf{Latent decomposition and mixture model.}
    To analyze the problem, we decompose each latent
    representation into a local temporal mean and a complementary centered residual:
    \begin{equation}
        \bar{\mathbf{z}}_t
        \triangleq
        \frac{1}{|\mathcal{W}_t|}
        \sum_{s\in\mathcal{W}_t}\mathbf{z}_s,
        \qquad
        \mathbf{r}_t
        \triangleq
        \mathbf{z}_t-\bar{\mathbf{z}}_t,
        \qquad
        \mathbf{z}_t
        \equiv
        \bar{\mathbf{z}}_t+\mathbf{r}_t,
        \label{eq:temporal_decomposition}
    \end{equation}
    where $\mathcal{W}_t$ denotes a local temporal window.
    We refer to $\bar{\mathbf{z}}_t$ as the temporally persistent component
    and $\mathbf{r}_t$ as the temporally centered residual. The persistent
    component captures variation that is shared within the local temporal window,
    while the centered residual captures complementary within-window variation.
    SIGReg applies the Epps--Pulley (EP) criterion to random one-dimensional marginal
    projections of the latent representation. For a projection vector
    $\mathbf{w}$, linearity preserves the temporal decomposition:
    \begin{equation}
        x
        =
        \mathbf{w}^{\top}\mathbf{z}_t
        =
        \underbrace{\mathbf{w}^{\top}\bar{\mathbf{z}}_t}_{c}
        +
        \underbrace{\mathbf{w}^{\top}\mathbf{r}_t}_{\epsilon}.
        \label{eq:projected_decomposition}
    \end{equation}
    We therefore study the Raw LeWM objective through the projected variables
    $c$ and $\epsilon$. For a controlled analysis, we model $c$ as a balanced
    $K$-component Gaussian mixture and
    $\epsilon\sim\mathcal{N}(0,\sigma_r^2)$ as Gaussian residual variation,
    with $\operatorname{Cov}(c,\epsilon)=0$. This yields
    \begin{equation}
        x = c + \varepsilon,
        \qquad
        c \sim \frac{1}{K}\sum_{k=1}^{K}
        \mathcal{N}\!\left(\mu_k,\ \tau^2\right),
        \qquad
        \varepsilon \sim \mathcal{N}\!\left(0,\ \sigma_r^2\right),
        \qquad
        \operatorname{Cov}(c,\varepsilon)=0,
        \label{eq:k_component_model}
    \end{equation}
    where $\mu_k$ is the projected center of the $k$-th component,
    $\tau$ is the common within-component standard deviation of the persistent
    component, and $\sigma_r$ is the scale of the residual variation.

    \noindent\textbf{Monte Carlo analysis of the Raw LeWM objective.}
    For each Monte Carlo layout, we sample and center $K$ component centers,
    normalize their variance to one, and rescale them by
    $\sqrt{\mathrm{var}(c)}$. The residual component is constructed analogously
    for a target $\mathrm{var}(\varepsilon)$. We set $\tau=0$, so that
    $\mathrm{var}(c)=\frac{1}{K}\sum_k\mu_k^2$ and
    $\mathrm{var}(\varepsilon)=\sigma_r^2$ are directly controlled and swept
    over a grid. We use $K=10$ and temporal windows of $W=4$, matching the
    suite-wise training setup. For each projected direction, we approximate the
    prediction cost as $\mathcal{L}_{\mathrm{pred}}=\mathrm{var}(\varepsilon)$
    and evaluate
    $
    \mathcal{L}_{\mathrm{Raw}}
    =
    \mathcal{L}_{\mathrm{pred}}
    +
    0.09\,\mathcal{L}_{\mathrm{EP}}(x).
    $
    Full details are provided in Appendix~\ref{app:mc_details}.

    \noindent\textbf{Results.}
    Figure~\ref{fig:mechanism}(a) shows
    $\mathcal{L}_{\mathrm{Raw}}$ over the
    $(\mathrm{var}(c),\mathrm{var}(\varepsilon))$ plane, with a clear minimum
    around $\mathrm{var}(c)\approx0.6$ and
    $\mathrm{var}(\varepsilon)\approx0.2$.
    Along each projection direction, the EP criterion encourages the projected
    marginal toward a standard Gaussian and therefore favors variance near one.
    Under the zero-cross-covariance setting used in our controlled analysis,
    $
    \mathrm{var}(x)=\mathrm{var}(c)+\mathrm{var}(\varepsilon),
    $
    so the persistent component and centered residual compete for the variance
    required by the projected marginal. Meanwhile, the prediction objective favors
    smaller residual variation. Their interaction therefore biases variance allocation
    toward the persistent component at the expense of the centered residual,
    suppressing residual variance.
    
    \noindent\textbf{Implication for policy learning.}
    The centered residual captures complementary within-window variation that
    can encode fine-grained changes in robot and environment states, including
    gripper motion, that are crucial for closed-loop manipulation.
    Suppressing this variation may therefore reduce
    the accessibility of robot state and dynamics from the learned representation
    and degrade downstream control, even without representation collapse.
    Our experiments in Sec.~\ref{sec:experiments} show that Raw LeWM
    representations have lower decodability of robot state and dynamics,
    particularly gripper dynamics, accompanied by poorer closed-loop
    manipulation performance.

    \section{Method}
    \label{sec:method}
    We introduce Temporally Centered LeWorldModel (TC-LeWM), which applies
    SIGReg to temporally centered residuals rather than directly to the original
    latent representations. This modification decouples persistent and
    residual variance allocation while retaining an effective anti-collapse
    property.

    \subsection{Temporally Centered LeWorldModel}

    \noindent\textbf{Model and training objective.}
    As illustrated in Fig.~\ref{fig:method_overview}, TC-LeWM retains the
    LeWM encoder and action-conditioned latent predictor,
    $\mathbf{z}_t=f_\theta(\mathbf{x}_t)$ and
    $\hat{\mathbf{z}}_{t+1}=g_\psi(\mathbf{z}_t,\mathbf{a}_t)$.
    For each latent sequence, we compute
    $
        \bar{\mathbf{z}}_t
        =
        \frac{1}{|\mathcal{W}_t|}
        \sum_{s\in\mathcal{W}_t}\mathbf{z}_s,
        \qquad
        \mathbf{r}_t
        =
        \mathbf{z}_t-\bar{\mathbf{z}}_t,
    $
    where $\mathcal{W}_t$ is a local temporal window. Let
    $\mathcal{R}=\{\mathbf{r}_t\}$ denote the residuals in a training batch.
    TC-LeWM optimizes
    \begin{equation}
        \mathcal{L}_{\mathrm{TC}}
        =
        \mathcal{L}_{\mathrm{pred}}
        +
        \lambda\,\mathcal{L}_{\mathrm{SIGReg}}(\mathcal{R}).
        \label{eq:tc_lewm_objective}
    \end{equation}
    Thus, the only change from Raw LeWM is the regularization target:
    SIGReg is applied to $\mathcal{R}$ instead of the original latent
    representations $\mathcal{Z}$.
    Further architectural and behavior-cloning implementation details are
    provided in Appendix~\ref{app:implementation}.

    \subsection{Why the Residual Target Works}

    \begin{table*}[t]
        \centering
        \caption{
            Per-task closed-loop success rates (\%) under \textbf{independent single-task
            training} on the ten LIBERO-Long tasks.
            Results are averaged over three training seeds and three evaluation seeds;
            standard deviations are omitted for space.
        }
        \label{tab:single_task}
        \setlength{\tabcolsep}{3.4pt}
        \resizebox{\textwidth}{!}{%
            \begin{tabular}{lcccccccccc|c}
                \toprule
                Method & T0 & T1 & T2 & T3 & T4 & T5 & T6 & T7 & T8 & T9
                & Avg. \\
                \midrule
                Raw LeWM
                & 53.1 & 37.5 & 12.4 & 11.8 & 11.6 & 13.6 & 29.8 & 28.7
                & 17.8 & 44.7 & 26.1 \\
                TC-LeWM (Ours)
                & \textbf{85.1} & \textbf{68.5} & \textbf{75.6}
                & \textbf{25.6} & \textbf{36.2} & \textbf{37.8}
                & \textbf{42.9} & \textbf{39.1} & \textbf{34.9}
                & \textbf{55.3} & \textbf{50.1} \\
                \midrule
                $\Delta$ (TC $-$ Raw)
                & +32.0 & +31.0 & +63.2 & +13.8 & +24.6 & +24.2 & +13.1
                & +10.4 & +17.1 & +10.6 & $\mathbf{+24.0}$ \\
                \bottomrule
            \end{tabular}%
        }
    \end{table*}

    \begin{table}[t]
        \centering
        \caption{
            Closed-loop success rates (\%) under \textbf{suite-wise 10-task training}.
            Results for Diffusion Policy, Octo, and OpenVLA are taken from
            \citet{kim24openvla}, while the $\pi_{0.5}$ results are taken from
            \citet{black2025pi05}.
            Raw LeWM and TC-LeWM are reported as
            mean$\pm$std over three full-pipeline training seeds.
        }
        \label{tab:suite10}
        \small
        \setlength{\tabcolsep}{3.8pt}
        \begin{tabular}{lcccc|c}
            \toprule
            Method & Spatial & Object & Goal & Long & Avg. \\
            \midrule

            \multicolumn{6}{c}{\textit{Pre-trained methods}} \\
            Octo (fine-tuned)
            & 78.9 & 85.7 & 84.6 & 51.1 & 75.1 \\
            OpenVLA (fine-tuned)
            & 84.7 & 88.4 & 79.2 & 53.7 & 76.5 \\
            $\pi_{0.5}$ (fine-tuned)
            & 98.8 & 98.2 & 98.0 & 92.4 & 96.9 \\

            \midrule
            \multicolumn{6}{c}{\textit{From-scratch methods}} \\
            Diffusion Policy
            & 78.3 & 92.5 & 68.3 & 50.5 & 72.4 \\

            \addlinespace[2pt]

            Raw LeWM
            & 62.4\std{7.5} & 93.9\std{2.3} & 53.4\std{5.3}
            & 44.7\std{4.3} & 63.6 \\
            TC-LeWM (Ours)
            & \textbf{86.0}\std{4.1} & \textbf{97.0}\std{0.1}
            & \textbf{88.4}\std{7.5} & \textbf{63.6}\std{6.8}
            & \textbf{83.8} \\
            \bottomrule
        \end{tabular}
    \end{table}

    Applying SIGReg to the temporally centered residual has two key properties:
    it decouples persistent and residual variance allocation and retains an
    effective anti-collapse property.

    \noindent\textbf{Decoupling persistent and residual variance allocation.}
    In Raw LeWM, the EP criterion is evaluated on the projected latent
    $x=c+\varepsilon$. Under the zero-cross-covariance setting in
    Sec.~\ref{sec:mc_analysis},
    $
        \mathrm{var}(x)
        =
        \mathrm{var}(c)
        +
        \mathrm{var}(\varepsilon),
    $
    so the persistent component and centered residual share the near-unit
    projected variance budget induced by SIGReg.
    TC-LeWM instead evaluates the EP criterion on the projected residual
    $\varepsilon=\mathbf{w}^{\top}\mathbf{r}_t$, which does not directly contain
    the persistent component $c$. Figure~\ref{fig:mechanism}(b) illustrates this
    difference: as $\mathrm{var}(c)$ increases,
    $\Delta\mathcal{L}_{\mathrm{EP}}$ increases for $x=c+\varepsilon$ but remains
    identically zero for $\varepsilon$. Thus, the residual remains regularized
    toward its standard-Gaussian target, while persistent variation no longer
    trades off against residual variance to satisfy the same projected target,
    decoupling their variance allocation.

    \noindent\textbf{Anti-collapse property retained.}
    If the representation collapses within a temporal window,
    $\mathbf{z}_t=\bar{\mathbf{z}}_t$ and therefore
    $\mathbf{r}_t=\mathbf{0}$.
    All one-dimensional projections of $\mathcal{R}$ then degenerate at zero and
    are strongly penalized by the EP criterion relative to the standard-Gaussian
    target. TC-LeWM therefore retains an effective anti-collapse property.
    The prediction and residual-regularization objectives remain complementary:
    unstructured residual variation may help satisfy SIGReg but is difficult to
    predict and is penalized by $\mathcal{L}_{\mathrm{pred}}$. Their combination
    therefore favors non-degenerate, predictable temporal variation without the
    persistent--residual coupling of Raw LeWM.

    \section{Experiments}
    \label{sec:experiments}

    \begin{table*}[t]
        \centering
        \caption{
            Closed-loop success rates (\%) under \textbf{unified 40-task world model
            training}. All four LIBERO suites share one encoder and latent predictor,
            while a separate task-conditioned downstream policy is trained for each
            suite. Results are mean$\pm$standard deviation over three full-pipeline
            training seeds, each evaluated using three evaluation seeds with
            50 rollouts per task.
        }
        \label{tab:libero40}
        \setlength{\tabcolsep}{5pt}
        \begin{tabular}{lcccc|c}
            \toprule
            Method & Spatial & Object & Goal & Long & Avg. \\
            \midrule
            Raw LeWM
            & 62.1\std{11.1} & 95.1\std{1.0} & 81.0\std{4.7}
            & 52.0\std{4.3} & 72.6 \\
            TC-LeWM (Ours)
            & \textbf{92.5}\std{3.0} & \textbf{97.1}\std{1.3}
            & \textbf{89.7}\std{2.5} & \textbf{62.3}\std{4.3}
            & \textbf{85.4} \\
            \midrule
            $\Delta$ (TC $-$ Raw)
            & $+30.4$ & $+2.0$ & $+8.7$ & $+10.3$
            & $\mathbf{+12.8}$ \\
            \bottomrule
        \end{tabular}
    \end{table*}

    We evaluate TC-LeWM on the LIBERO benchmark through closed-loop manipulation
    and analyze how temporal centering changes the learned representations.

    \subsection{Experimental Setup}
    \label{sec:experimental_setup}

    \noindent\textbf{Benchmark.}
    LIBERO~\citep{liu2023libero} contains four suites---Spatial, Object, Goal, and Long---each consisting
    of ten manipulation tasks. Each observation contains two RGB camera views.
    A shared ViT encoder \citep{dosovitskiy2021an} processes the two views
    independently. We reconstruct the LIBERO datasets following OpenVLA \citep{kim24openvla}.

    \noindent\textbf{World model training and downstream policy learning.}
    We study downstream robot policy learning using a two-stage pipeline.
    First, the encoder and action-conditioned latent predictor are trained with
    the world model objective on reward-free demonstration trajectories.
    At each frame, the CLS tokens from the two camera views are concatenated
    and passed through a shared projector to obtain the latent representation
    regularized by SIGReg, while patch tokens are not directly regularized.
    After world model training, the encoder is frozen and a task-conditioned
    flow-matching behavior-cloning policy is trained on the frozen visual
    representations. For downstream policy learning, we use both the CLS and
    patch tokens from the two camera views.

    This setting allows us to examine how representations learned by the world
    model affect downstream robot policy performance.
    Raw LeWM and TC-LeWM use identical frozen-encoder and downstream-policy
    settings.

    \noindent\textbf{Task aggregation settings.}
    We consider three levels of task aggregation.
    In the \textbf{single-task} setting, one world model and one downstream policy
    are trained independently for each LIBERO-Long task.
    In the \textbf{10-task} setting, the ten tasks within each suite share one
    encoder, latent predictor, and task-conditioned downstream policy.
    In the \textbf{40-task} setting, data from all four suites are aggregated to
    train one unified world model. After freezing the encoder, we train a separate
    task-conditioned downstream policy for each suite, such that each downstream
    policy still covers ten tasks as in the 10-task setting. This design isolates
    the effect of scaling the representation learner from jointly training a single 40-task action policy.

    \noindent\textbf{Compared methods.}
    We mainly compare \textbf{Raw LeWM}, which applies SIGReg to the latent
    representations $\mathbf{z}_t$, with \textbf{TC-LeWM}, which applies the same
    regularizer to the temporally centered residual
    $\mathbf{r}_t=\mathbf{z}_t-\bar{\mathbf{z}}_t$.
    Raw LeWM and TC-LeWM share the same architecture, training data, and downstream policy protocol. 
    TC-LeWM uses a temporal window of $W=4$ frames.

    \noindent\textbf{Closed-loop evaluation.}
    Each model is evaluated using three
    evaluation seeds with 50 rollouts per task.
    We report closed-loop task success rates as mean$\pm$standard deviation over
    three full-pipeline training seeds.

    \subsection{Closed-Loop Manipulation Performance}
    \label{sec:single_task}

    \noindent\textbf{Single-task training.}
    Table~\ref{tab:single_task} shows that when each LIBERO-Long task is trained
    independently, the downstream policy using Raw LeWM representations achieves
    only $26.1\%$ average success, whereas the policy using TC-LeWM
    representations reaches $50.1\%$.
    TC-LeWM improves all ten tasks, yielding an
    average gain of $+24.0$ percentage points. These results show that avoiding
    representation collapse alone does not guarantee representations suitable for
    downstream manipulation policy learning.

    \noindent\textbf{Suite-wise 10-task training.}
    We next jointly train the ten tasks within each LIBERO suite, following the
    standard suite-wise evaluation setting of LIBERO.
    As shown in Table~\ref{tab:suite10}, TC-LeWM consistently outperforms Raw LeWM
    across all four suites, raising the average success rate from $63.6\%$ to
    $83.8\%$, with particularly large gains on LIBERO-Goal and LIBERO-Spatial.
    Among methods trained from scratch, TC-LeWM achieves the highest average
    success rate, outperforming Diffusion Policy~\citep{chi2023diffusionpolicy}.
    It also surpasses the pretrained Octo~\citep{octo_2023} and OpenVLA
    baselines, while large-scale pretrained $\pi_{0.5}$~\citep{black2025pi05}
    remains stronger.

    \noindent\textbf{Unified 40-task world model training.}
    We further aggregate all 40 LIBERO tasks to train a single world model.
    Table~\ref{tab:libero40} shows that TC-LeWM again outperforms Raw LeWM on
    all four suites, achieving $85.4\%$ average success compared with $72.6\%$
    for Raw LeWM, an absolute improvement of $12.8$ percentage points.
    The largest gain is observed on LIBERO-Spatial, where TC-LeWM improves
    success from $62.1\%$ to $92.5\%$, while substantial gains also remain on
    LIBERO-Goal and LIBERO-Long. Raw LeWM benefits from the larger training set, most notably on
    LIBERO-Goal, but the absolute performance advantage of TC-LeWM persists
    even under unified 40-task world model training. 
    This indicates that scaling the world-model training data alone does not
    remove the representation gap addressed by temporal centering.

    \subsection{Analysis of Learned Representations}
    \label{sec:latent_analysis}

    \begin{figure*}[t]
        \centering
        \includegraphics[width=\textwidth]{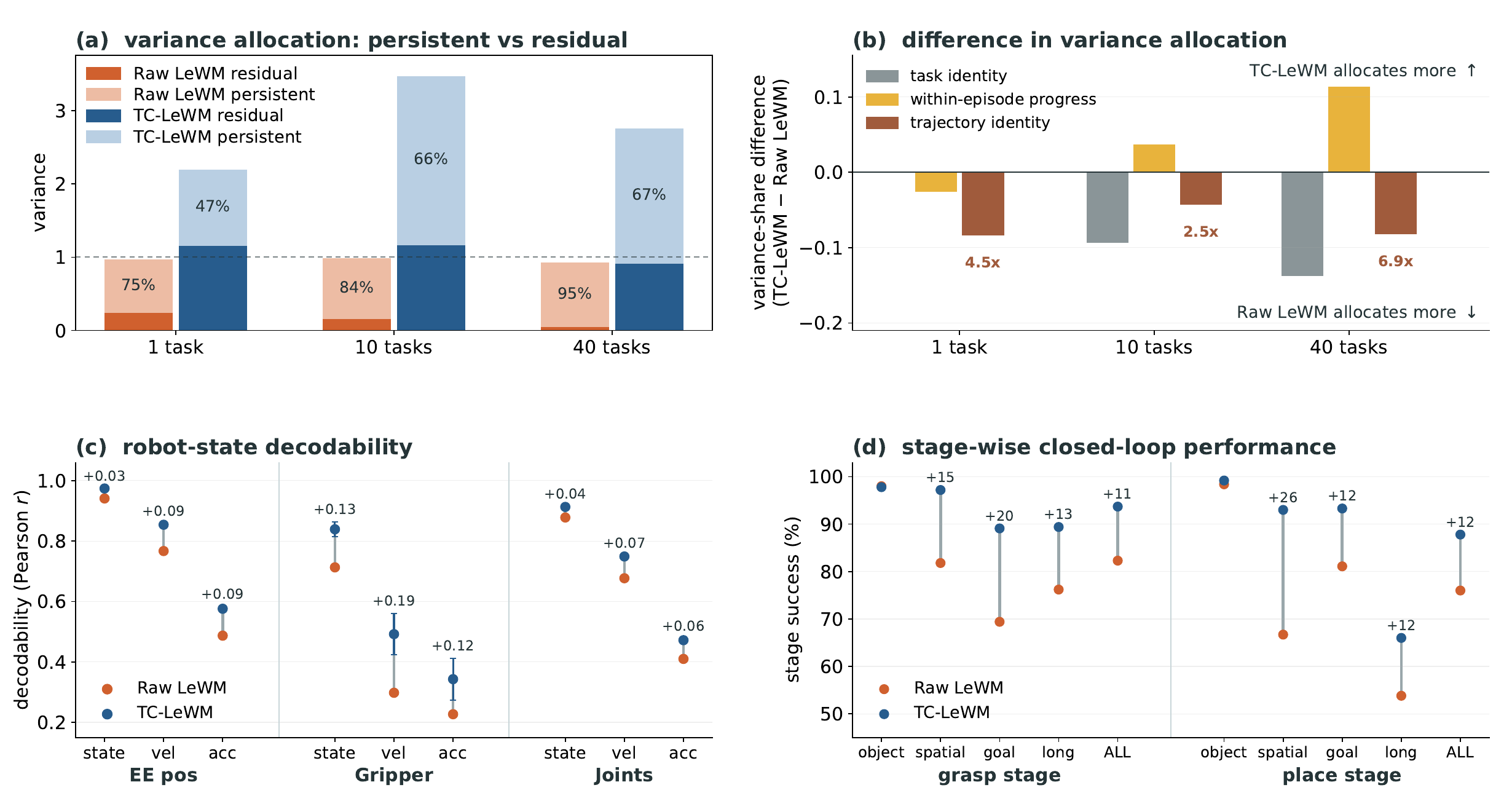}
        \caption{
            \textbf{Analysis of learned representations}, averaged over three training seeds.
            \textbf{(a)} Variance allocation between the temporally persistent
            component and centered residual across training scales.
            \textbf{(b)} Variance associated with task identity, within-episode
            progress, and trajectory identity.
            \textbf{(c)} Linear decodability of end-effector, gripper, and joint
            state and dynamics.
            \textbf{(d)} Closed-loop success at the grasping and placement stages.
        }
        \label{fig:analysis}
    \end{figure*}

    We next analyze how differences in the learned representations are reflected
    in encoded information, robot-state decodability, and closed-loop behavior.

    \noindent\textbf{Variance allocation.}
    Figure~\ref{fig:mechanism}(c) directly measures
    $\mathrm{var}(\bar{\mathbf z})$ and $\mathrm{var}(\mathbf r)$ in trained
    10-task representations. Raw LeWM exhibits a nearly fixed total variance
    dominated by the persistent component, consistent with the coupled allocation
    predicted by the Monte Carlo analysis. In contrast, TC-LeWM maintains a stable
    residual variance while allowing persistent variance to vary across suites,
    consistent with decoupled variance allocation.

    Figure~\ref{fig:analysis}(a) further shows that this distinction persists
    across all three training scales. In Raw LeWM, the temporally persistent
    component dominates the representation, while the centered residual accounts
    for less than approximately $25\%$ of the total variance. TC-LeWM instead
    maintains substantial variation in both components: the persistent component
    accounts for approximately $47$--$67\%$ of the total variance, while the
    centered residual retains a substantial share at every scale. Thus, neither
    component consistently dominates the learned representation under TC-LeWM.

    \noindent\textbf{Content attribution.}
    Figure~\ref{fig:analysis}(b) analyzes how representation variance is associated
    with task identity, within-episode progress, and trajectory identity.
    At the multi-task scales, Raw LeWM allocates more variance to task identity
    than TC-LeWM, and across all training scales it allocates
    $2.5$--$6.9\times$ more variance to trajectory identity.
    Task identity is partly redundant because the downstream policy already
    receives the task label explicitly, while trajectory identity largely
    captures trajectory-specific variation that is less relevant
    to fine-grained manipulation.

    In contrast, TC-LeWM allocates relatively more variance to within-episode
    progress under multi-task training. 
    Such variation reflects the temporal
    evolution of the manipulation trajectory and is more directly aligned with
    the changing state required for closed-loop control. These results suggest
    that TC-LeWM not only preserves more residual variation, but also shifts the
    representation away from identity-specific variation toward temporally
    structured information.

    \begin{figure*}[t]
        \centering
        \includegraphics[width=\textwidth]{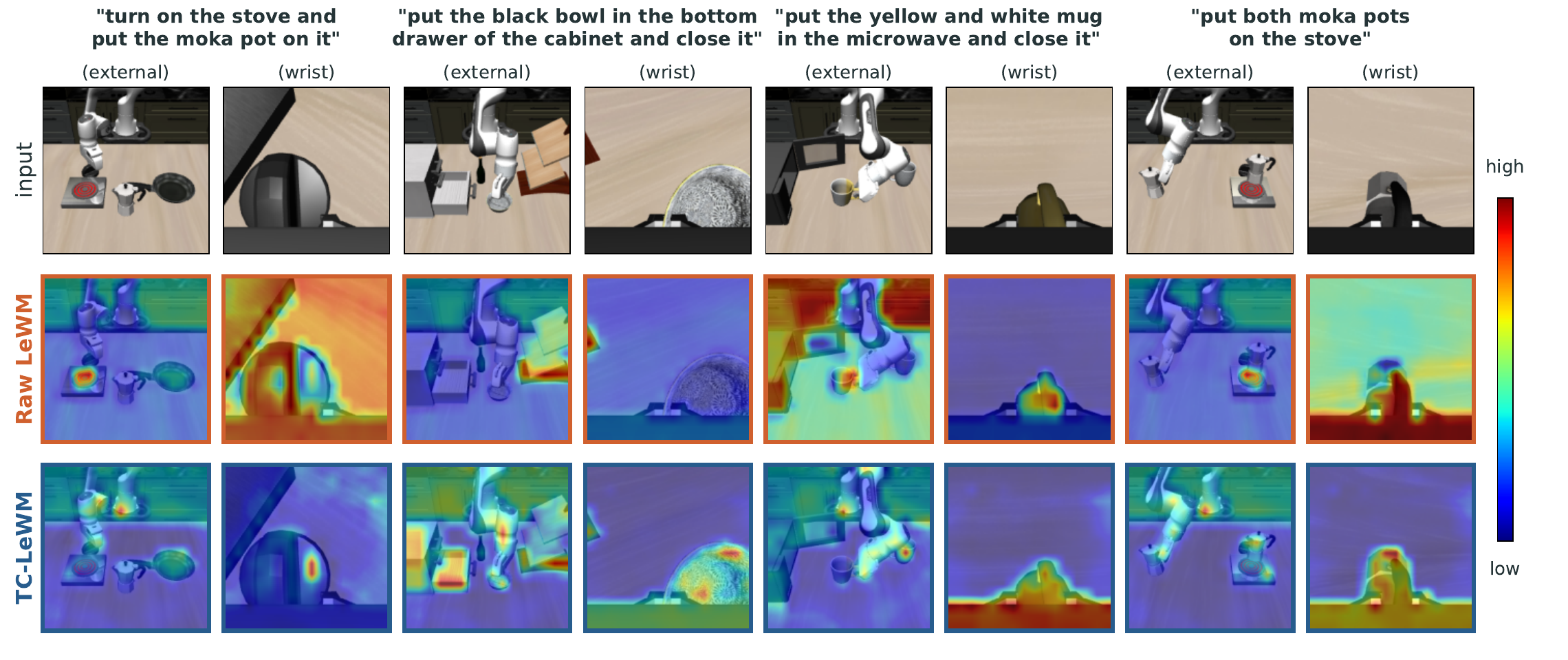}
        \caption{
            \textbf{Attention-rollout visualization at the grasping moment} for the first
            four LIBERO-Long tasks. TC-LeWM more consistently attends to the robot
            arm, gripper, and manipulated object, whereas Raw LeWM more frequently
            attends to background or task-irrelevant regions.
        }
        \label{fig:attn}
    \end{figure*}

    \noindent\textbf{Decodability of robot state and dynamics.}
    We next test whether the preserved temporal variation makes robot state and
    dynamics more accessible from the learned representation.
    Figure~\ref{fig:analysis}(c) reports linear decodability from the 10-task
    representations, averaged over three training seeds, for end-effector,
    gripper, and joint quantities at the state, velocity, and acceleration
    levels. TC-LeWM achieves higher decodability than Raw LeWM for all three
    quantities and at all derivative orders. The advantage is particularly
    pronounced for gripper dynamics and generally becomes larger for higher-order
    temporal information.
    Reduced centered-residual variation in Raw LeWM
    is accompanied by lower accessibility of fine-grained robot state and
    dynamics, whereas TC-LeWM preserves these quantities more effectively.

    \noindent\textbf{Behavioral analysis.}
    Given the improved decodability of robot state and dynamics, we further
    examine where the downstream performance gap emerges during closed-loop
    execution. Figure~\ref{fig:analysis}(d) decomposes manipulation into grasping
    and placement stages. TC-LeWM achieves substantially higher success than
    Raw LeWM at both stages.

    Both grasping and placement require fine-grained control of the robot arm
    and depend strongly on the evolving robot state and dynamics. The consistent
    performance advantage across both stages is therefore aligned with the
    improved decodability of robot state and dynamics from TC-LeWM
    representations observed in Fig.~\ref{fig:analysis}(c).

    \noindent\textbf{Attention visualization.}
    Finally, we qualitatively examine the visual features used by the two encoders
    through attention rollout \citep{abnar2020quantifying} at the grasping moment.
    As shown in Fig.~\ref{fig:attn}, TC-LeWM more consistently attends to
    manipulation-relevant regions, including the robot arm, gripper, and
    manipulated object, whereas Raw LeWM more often attends to background or
    task-irrelevant regions. The same tendency appears in the wrist-camera view,
    where TC-LeWM more consistently attends jointly to the gripper and grasped
    object. These observations provide complementary evidence that TC-LeWM learns
    visual representations more closely aligned with robot-object interaction.
    Implementation details and additional latent-space analyses are provided in
    Appendices~\ref{app:representation_details}
    and~\ref{app:additional_latent_analysis}.

    \section{Conclusion}
    In this work, we identified a variance-allocation bias in the Raw LeWM objective:
    the interaction between latent prediction and SIGReg favors temporally
    persistent variation while suppressing the centered residual.
    Based on this analysis, we introduced TC-LeWM, which applies SIGReg to
    temporally centered residuals to decouple their variance allocation while
    retaining an effective anti-collapse property.
    Across single-task, suite-wise 10-task, and unified 40-task LIBERO settings,
    TC-LeWM consistently improves the closed-loop performance of downstream
    behavior-cloning policies and yields representations with more linearly
    decodable robot state and dynamics, particularly gripper dynamics.
    Future work will scale TC-LeWM to larger reward-free datasets and explore its
    integration with Vision-Language-Action and World-Action models for
    open-world manipulation.

\newpage
\section*{Acknowledgments}
    This work was supported in part by JST PRESTO, Japan, under Grant JPMJPR25T5.

\bibliography{main}

\begin{thebibliography}{37}
\providecommand{\natexlab}[1]{#1}
\providecommand{\url}[1]{\texttt{#1}}
\expandafter\ifx\csname urlstyle\endcsname\relax
  \providecommand{\doi}[1]{doi: #1}\else
  \providecommand{\doi}{doi: \begingroup \urlstyle{rm}\Url}\fi

\bibitem[Abnar \& Zuidema(2020)Abnar and Zuidema]{abnar2020quantifying}
Samira Abnar and Willem Zuidema.
\newblock Quantifying attention flow in transformers.
\newblock In \emph{Proceedings of the 58th annual meeting of the association for computational linguistics}, pp.\  4190--4197, 2020.

\bibitem[Assran et~al.(2023)Assran, Duval, Misra, Bojanowski, Vincent, Rabbat, LeCun, and Ballas]{assran2023self}
Mahmoud Assran, Quentin Duval, Ishan Misra, Piotr Bojanowski, Pascal Vincent, Michael Rabbat, Yann LeCun, and Nicolas Ballas.
\newblock Self-supervised learning from images with a joint-embedding predictive architecture.
\newblock In \emph{Proceedings of the IEEE/CVF conference on computer vision and pattern recognition}, pp.\  15619--15629, 2023.

\bibitem[Assran et~al.(2025)Assran, Bardes, Fan, Garrido, Howes, Komeili, Muckley, Rizvi, Roberts, Sinha, Zholus, Arnaud, Gejji, Martin, Hogan, Dugas, Bojanowski, Khalidov, Labatut, Massa, Szafraniec, Krishnakumar, Li, Ma, Chandar, Meier, LeCun, Rabbat, and Ballas]{assran2025v}
Mido Assran, Adrien Bardes, David Fan, Quentin Garrido, Russell Howes, Mojtaba Komeili, Matthew Muckley, Ammar Rizvi, Claire Roberts, Koustuv Sinha, Artem Zholus, Sergio Arnaud, Abha Gejji, Ada Martin, Francois~Robert Hogan, Daniel Dugas, Piotr Bojanowski, Vasil Khalidov, Patrick Labatut, Francisco Massa, Marc Szafraniec, Kapil Krishnakumar, Yong Li, Xiaodong Ma, Sarath Chandar, Franziska Meier, Yann LeCun, Michael Rabbat, and Nicolas Ballas.
\newblock {V-JEPA 2}: Self-supervised video models enable understanding, prediction and planning.
\newblock \emph{arXiv preprint arXiv:2506.09985}, 2025.

\bibitem[Balestriero \& LeCun(2024)Balestriero and LeCun]{balestriero2024learning}
Randall Balestriero and Yann LeCun.
\newblock How learning by reconstruction produces uninformative features for perception.
\newblock In \emph{Proceedings of the 41st International Conference on Machine Learning}, volume 235 of \emph{Proceedings of Machine Learning Research}, pp.\  2566--2585. PMLR, 2024.

\bibitem[Balestriero \& LeCun(2025)Balestriero and LeCun]{balestriero2025lejepa}
Randall Balestriero and Yann LeCun.
\newblock {LeJEPA}: Provable and scalable self-supervised learning without the heuristics.
\newblock \emph{arXiv preprint arXiv:2511.08544}, 2025.

\bibitem[Bardes et~al.(2022)Bardes, Ponce, and LeCun]{Bardes2021VICRegVR}
Adrien Bardes, Jean Ponce, and Yann LeCun.
\newblock {VICReg}: Variance-invariance-covariance regularization for self-supervised learning.
\newblock In \emph{International Conference on Learning Representations}, 2022.

\bibitem[Bardes et~al.(2024)Bardes, Garrido, Ponce, Chen, Rabbat, LeCun, Assran, and Ballas]{Bardes2024RevisitingFP}
Adrien Bardes, Quentin Garrido, Jean Ponce, Xinlei Chen, Michael Rabbat, Yann LeCun, Mido Assran, and Nicolas Ballas.
\newblock Revisiting feature prediction for learning visual representations from video.
\newblock \emph{Transactions on Machine Learning Research}, 2024.
\newblock ISSN 2835-8856.
\newblock Featured Certification.

\bibitem[Black et~al.(2025)Black, Brown, Darpinian, Dhabalia, Driess, Esmail, Equi, Finn, Fusai, Galliker, Ghosh, Groom, Hausman, Ichter, Jakubczak, Jones, Ke, LeBlanc, Levine, Li-Bell, Mothukuri, Nair, Pertsch, Ren, Shi, Smith, Springenberg, Stachowicz, Tanner, Vuong, Walke, Walling, Wang, Yu, and Zhilinsky]{black2025pi05}
Kevin Black, Noah Brown, James Darpinian, Karan Dhabalia, Danny Driess, Adnan Esmail, Michael~Robert Equi, Chelsea Finn, Niccolo Fusai, Manuel~Y. Galliker, Dibya Ghosh, Lachy Groom, Karol Hausman, Brian Ichter, Szymon Jakubczak, Tim Jones, Liyiming Ke, Devin LeBlanc, Sergey Levine, Adrian Li-Bell, Mohith Mothukuri, Suraj Nair, Karl Pertsch, Allen~Z. Ren, Lucy~Xiaoyang Shi, Laura Smith, Jost~Tobias Springenberg, Kyle Stachowicz, James Tanner, Quan Vuong, Homer Walke, Anna Walling, Haohuan Wang, Lili Yu, and Ury Zhilinsky.
\newblock $\pi_{0.5}$: A vision-language-action model with open-world generalization.
\newblock In Joseph Lim, Shuran Song, and Hae-Won Park (eds.), \emph{Proceedings of The 9th Conference on Robot Learning}, volume 305 of \emph{Proceedings of Machine Learning Research}, pp.\  17--40. PMLR, 2025.

\bibitem[Chen \& He(2021)Chen and He]{Chen2020ExploringSS}
Xinlei Chen and Kaiming He.
\newblock Exploring simple siamese representation learning.
\newblock In \emph{Proceedings of the IEEE/CVF Conference on Computer Vision and Pattern Recognition}, pp.\  15750--15758, 2021.

\bibitem[Chi et~al.(2023)Chi, Feng, Du, Xu, Cousineau, Burchfiel, and Song]{chi2023diffusionpolicy}
Cheng Chi, Siyuan Feng, Yilun Du, Zhenjia Xu, Eric Cousineau, Benjamin C.~M. Burchfiel, and Shuran Song.
\newblock Diffusion policy: Visuomotor policy learning via action diffusion.
\newblock In \emph{Proceedings of Robotics: Science and Systems}, Daegu, Republic of Korea, July 2023.
\newblock \doi{10.15607/RSS.2023.XIX.026}.

\bibitem[Dawid \& LeCun(2024)Dawid and LeCun]{dawid2024introduction}
Anna Dawid and Yann LeCun.
\newblock Introduction to latent variable energy-based models: A path toward autonomous machine intelligence.
\newblock \emph{Journal of Statistical Mechanics: Theory and Experiment}, 2024\penalty0 (10):\penalty0 104011, 2024.
\newblock \doi{10.1088/1742-5468/ad292b}.

\bibitem[Deng et~al.(2022)Deng, Jang, and Ahn]{Deng2021DreamerProRM}
Fei Deng, Ingook Jang, and Sungjin Ahn.
\newblock {DreamerPro}: Reconstruction-free model-based reinforcement learning with prototypical representations.
\newblock In \emph{Proceedings of the 39th International Conference on Machine Learning}, volume 162 of \emph{Proceedings of Machine Learning Research}, pp.\  4956--4975. PMLR, 2022.

\bibitem[Dosovitskiy et~al.(2021)Dosovitskiy, Beyer, Kolesnikov, Weissenborn, Zhai, Unterthiner, Dehghani, Minderer, Heigold, Gelly, Uszkoreit, and Houlsby]{dosovitskiy2021an}
Alexey Dosovitskiy, Lucas Beyer, Alexander Kolesnikov, Dirk Weissenborn, Xiaohua Zhai, Thomas Unterthiner, Mostafa Dehghani, Matthias Minderer, Georg Heigold, Sylvain Gelly, Jakob Uszkoreit, and Neil Houlsby.
\newblock An image is worth 16x16 words: Transformers for image recognition at scale.
\newblock In \emph{International Conference on Learning Representations}, 2021.

\bibitem[Epps \& Pulley(1983)Epps and Pulley]{epps1983test}
T.~W. Epps and Lawrence~B. Pulley.
\newblock A test for normality based on the empirical characteristic function.
\newblock \emph{Biometrika}, 70\penalty0 (3):\penalty0 723--726, 1983.
\newblock \doi{10.1093/biomet/70.3.723}.

\bibitem[Fu et~al.(2021)Fu, Yang, Agrawal, and Jaakkola]{Fu2021LearningTI}
Xiang Fu, Ge~Yang, Pulkit Agrawal, and Tommi Jaakkola.
\newblock Learning task informed abstractions.
\newblock In \emph{Proceedings of the 38th International Conference on Machine Learning}, volume 139 of \emph{Proceedings of Machine Learning Research}, pp.\  3480--3491. PMLR, 2021.

\bibitem[Ghosh et~al.(2024)Ghosh, Walke, Pertsch, Black, Mees, Dasari, Hejna, Kreiman, Xu, Luo, Tan, Chen, Vuong, Xiao, Sanketi, Sadigh, Finn, and Levine]{octo_2023}
Dibya Ghosh, Homer~Rich Walke, Karl Pertsch, Kevin Black, Oier Mees, Sudeep Dasari, Joey Hejna, Tobias Kreiman, Charles Xu, Jianlan Luo, You~Liang Tan, Lawrence~Yunliang Chen, Quan Vuong, Ted Xiao, Pannag~R. Sanketi, Dorsa Sadigh, Chelsea Finn, and Sergey Levine.
\newblock {Octo}: An open-source generalist robot policy.
\newblock In \emph{Proceedings of Robotics: Science and Systems}, Delft, Netherlands, July 2024.
\newblock \doi{10.15607/RSS.2024.XX.090}.

\bibitem[Grill et~al.(2020)Grill, Strub, Altch{\'e}, Tallec, Richemond, Buchatskaya, Doersch, Avila~Pires, Guo, Azar, Piot, Kavukcuoglu, Munos, and Valko]{Grill2020BootstrapYO}
Jean-Bastien Grill, Florian Strub, Florent Altch{\'e}, Corentin Tallec, Pierre~H. Richemond, Elena Buchatskaya, Carl Doersch, Bernardo Avila~Pires, Zhaohan~Daniel Guo, Mohammad~Gheshlaghi Azar, Bilal Piot, Koray Kavukcuoglu, R{\'e}mi Munos, and Michal Valko.
\newblock Bootstrap your own latent - a new approach to self-supervised learning.
\newblock In \emph{Advances in Neural Information Processing Systems}, volume~33, pp.\  21271--21284, 2020.

\bibitem[Ha \& Schmidhuber(2018)Ha and Schmidhuber]{ha2018world}
David Ha and J{\"u}rgen Schmidhuber.
\newblock Recurrent world models facilitate policy evolution.
\newblock In \emph{Advances in Neural Information Processing Systems}, volume~31, pp.\  2455--2467, 2018.

\bibitem[Hafner et~al.(2019)Hafner, Lillicrap, Fischer, Villegas, Ha, Lee, and Davidson]{hafner2019learning}
Danijar Hafner, Timothy Lillicrap, Ian Fischer, Ruben Villegas, David Ha, Honglak Lee, and James Davidson.
\newblock Learning latent dynamics for planning from pixels.
\newblock In \emph{Proceedings of the 36th International Conference on Machine Learning}, volume~97 of \emph{Proceedings of Machine Learning Research}, pp.\  2555--2565. PMLR, 2019.

\bibitem[Hafner et~al.(2020)Hafner, Lillicrap, Ba, and Norouzi]{hafner2019dream}
Danijar Hafner, Timothy Lillicrap, Jimmy Ba, and Mohammad Norouzi.
\newblock Dream to control: Learning behaviors by latent imagination.
\newblock In \emph{International Conference on Learning Representations}, 2020.

\bibitem[Hafner et~al.(2021)Hafner, Lillicrap, Norouzi, and Ba]{Hafner2020MasteringAW}
Danijar Hafner, Timothy~P. Lillicrap, Mohammad Norouzi, and Jimmy Ba.
\newblock Mastering atari with discrete world models.
\newblock In \emph{International Conference on Learning Representations}, 2021.

\bibitem[Hafner et~al.(2025)Hafner, Pasukonis, Ba, and Lillicrap]{hafner2023mastering}
Danijar Hafner, Jurgis Pasukonis, Jimmy Ba, and Timothy Lillicrap.
\newblock Mastering diverse control tasks through world models.
\newblock \emph{Nature}, 640:\penalty0 647--653, 2025.
\newblock \doi{10.1038/s41586-025-08744-2}.

\bibitem[Hutson et~al.(2024)Hutson, Kauvar, and Haber]{hutson2024policy}
Miles Hutson, Isaac Kauvar, and Nick Haber.
\newblock Policy-shaped prediction: Avoiding distractions in model-based reinforcement learning.
\newblock In \emph{Advances in Neural Information Processing Systems}, volume~37, pp.\  13124--13148, 2024.
\newblock \doi{10.52202/079017-0418}.

\bibitem[Kim et~al.(2025)Kim, Pertsch, Karamcheti, Xiao, Balakrishna, Nair, Rafailov, Foster, Sanketi, Vuong, Kollar, Burchfiel, Tedrake, Sadigh, Levine, Liang, and Finn]{kim24openvla}
Moo~Jin Kim, Karl Pertsch, Siddharth Karamcheti, Ted Xiao, Ashwin Balakrishna, Suraj Nair, Rafael Rafailov, Ethan~P. Foster, Pannag~R. Sanketi, Quan Vuong, Thomas Kollar, Benjamin Burchfiel, Russ Tedrake, Dorsa Sadigh, Sergey Levine, Percy Liang, and Chelsea Finn.
\newblock {OpenVLA}: An open-source vision-language-action model.
\newblock In \emph{Proceedings of The 8th Conference on Robot Learning}, volume 270 of \emph{Proceedings of Machine Learning Research}, pp.\  2679--2713. PMLR, 2025.

\bibitem[Lipman et~al.(2023)Lipman, Chen, Ben-Hamu, Nickel, and Le]{lipman2023flow}
Yaron Lipman, Ricky T.~Q. Chen, Heli Ben-Hamu, Maximilian Nickel, and Matthew Le.
\newblock Flow matching for generative modeling.
\newblock In \emph{International Conference on Learning Representations}, 2023.
\newblock URL \url{https://openreview.net/forum?id=PqvMRDCJT9t}.

\bibitem[Liu et~al.(2023)Liu, Zhu, Gao, Feng, Liu, Zhu, and Stone]{liu2023libero}
Bo~Liu, Yifeng Zhu, Chongkai Gao, Yihao Feng, Qiang Liu, Yuke Zhu, and Peter Stone.
\newblock Libero: Benchmarking knowledge transfer for lifelong robot learning.
\newblock \emph{Advances in Neural Information Processing Systems}, 36:\penalty0 44776--44791, 2023.

\bibitem[Maes et~al.(2026)Maes, Le~Lidec, Scieur, LeCun, and Balestriero]{maes2026leworldmodel}
Lucas Maes, Quentin Le~Lidec, Damien Scieur, Yann LeCun, and Randall Balestriero.
\newblock {LeWorldModel}: Stable end-to-end joint-embedding predictive architecture from pixels.
\newblock \emph{arXiv preprint arXiv:2603.19312}, 2026.

\bibitem[Peebles \& Xie(2023)Peebles and Xie]{peebles2023scalable}
William Peebles and Saining Xie.
\newblock Scalable diffusion models with transformers.
\newblock In \emph{Proceedings of the IEEE/CVF International Conference on Computer Vision}, pp.\  4195--4205, 2023.

\bibitem[Schrittwieser et~al.(2020)Schrittwieser, Antonoglou, Hubert, Simonyan, Sifre, Schmitt, Guez, Lockhart, Hassabis, Graepel, Lillicrap, and Silver]{schrittwieser2020mastering}
Julian Schrittwieser, Ioannis Antonoglou, Thomas Hubert, Karen Simonyan, Laurent Sifre, Simon Schmitt, Arthur Guez, Edward Lockhart, Demis Hassabis, Thore Graepel, Timothy Lillicrap, and David Silver.
\newblock Mastering atari, go, chess and shogi by planning with a learned model.
\newblock \emph{Nature}, 588\penalty0 (7839):\penalty0 604--609, 2020.
\newblock \doi{10.1038/s41586-020-03051-4}.

\bibitem[Sobal et~al.(2025)Sobal, Zhang, Cho, Balestriero, Rudner, and LeCun]{Sobal2025LearningFR}
Uladzislau Sobal, Wancong Zhang, Kyunghyun Cho, Randall Balestriero, Tim G.~J. Rudner, and Yann LeCun.
\newblock Learning from reward-free offline data: A case for planning with latent dynamics models.
\newblock In \emph{Advances in Neural Information Processing Systems}, volume~38, 2025.

\bibitem[Wan et~al.(2023)Wan, Wang, Shao, Chen, and Zhan]{Wan2023SeMAILED}
Shenghua Wan, Yucen Wang, Minghao Shao, Ruying Chen, and De-Chuan Zhan.
\newblock {SeMAIL}: Eliminating distractors in visual imitation via separated models.
\newblock In \emph{Proceedings of the 40th International Conference on Machine Learning}, volume 202 of \emph{Proceedings of Machine Learning Research}, pp.\  35426--35443. PMLR, 2023.

\bibitem[Wang et~al.(2024)Wang, Wan, Gan, Feng, and Zhan]{Wang2024AD3IA}
Yucen Wang, Shenghua Wan, Le~Gan, Shuai Feng, and De-Chuan Zhan.
\newblock {AD3}: Implicit action is the key for world models to distinguish the diverse visual distractors.
\newblock In \emph{Proceedings of the 41st International Conference on Machine Learning}, volume 235 of \emph{Proceedings of Machine Learning Research}, pp.\  51546--51568. PMLR, 2024.

\bibitem[Watter et~al.(2015)Watter, Springenberg, Boedecker, and Riedmiller]{watter2015embed}
Manuel Watter, Jost Springenberg, Joschka Boedecker, and Martin Riedmiller.
\newblock Embed to control: A locally linear latent dynamics model for control from raw images.
\newblock In \emph{Advances in Neural Information Processing Systems}, volume~28, pp.\  2746--2754, 2015.

\bibitem[Wu et~al.(2023)Wu, Escontrela, Hafner, Abbeel, and Goldberg]{wu2022daydreamer}
Philipp Wu, Alejandro Escontrela, Danijar Hafner, Pieter Abbeel, and Ken Goldberg.
\newblock Daydreamer: World models for physical robot learning.
\newblock In \emph{Proceedings of The 6th Conference on Robot Learning}, volume 205 of \emph{Proceedings of Machine Learning Research}, pp.\  2226--2240. PMLR, 2023.

\bibitem[Zbontar et~al.(2021)Zbontar, Jing, Misra, LeCun, and Deny]{Zbontar2021BarlowTS}
Jure Zbontar, Li~Jing, Ishan Misra, Yann LeCun, and St{\'e}phane Deny.
\newblock Barlow twins: Self-supervised learning via redundancy reduction.
\newblock In \emph{Proceedings of the 38th International Conference on Machine Learning}, volume 139 of \emph{Proceedings of Machine Learning Research}, pp.\  12310--12320. PMLR, 2021.

\bibitem[Zhou et~al.(2025)Zhou, Pan, LeCun, and Pinto]{Zhou2024DINOWMWM}
Gaoyue Zhou, Hengkai Pan, Yann LeCun, and Lerrel Pinto.
\newblock {DINO-WM}: World models on pre-trained visual features enable zero-shot planning.
\newblock In \emph{Proceedings of the 42nd International Conference on Machine Learning}, volume 267 of \emph{Proceedings of Machine Learning Research}, pp.\  79115--79135. PMLR, 2025.

\bibitem[Zhu et~al.(2023)Zhu, Simchowitz, Gadipudi, and Gupta]{Zhu2023RePoRM}
Chuning Zhu, Max Simchowitz, Siri Gadipudi, and Abhishek Gupta.
\newblock {RePo}: Resilient model-based reinforcement learning by regularizing posterior predictability.
\newblock In \emph{Advances in Neural Information Processing Systems}, volume~36, pp.\  32445--32467, 2023.

\end{thebibliography}
\bibliographystyle{conference}

\newpage
\appendix
\section{Appendix}
\label{sec:appendix}

\subsection{Model and Implementation Details}
    \label{app:implementation}

    \noindent\textbf{Multi-view extension.}
    We retain the encoder and latent predictor architecture of LeWM and extend
    its visual input to the two RGB views used in LIBERO: an external camera
    and a wrist-mounted camera. A parameter-shared ViT encoder processes the
    two views independently. Let
    \begin{equation}
        \mathbf h_t^{(k)}
        =
        \operatorname{CLS}
        \left(
            E_\theta(\mathbf x_t^{(k)})
        \right),
        \qquad
        k\in\{1,\ldots,K_{\mathrm{cam}}\},
    \end{equation}
    denote the CLS token of camera view $k$. The per-view CLS tokens are
    concatenated and passed through the LeWM projector to form a single latent
    representation:
    \begin{equation}
        \mathbf z_t
        =
        p_\theta
        \left(
            [
                \mathbf h_t^{(1)};
                \ldots;
                \mathbf h_t^{(K_{\mathrm{cam}})}
            ]
        \right).
        \label{eq:app_multiview}
    \end{equation}
    For LIBERO, $K_{\mathrm{cam}}=2$. Patch tokens are not directly regularized
    by SIGReg but are retained for downstream policy learning. Apart from this
    multi-view extension, we keep the LeWM encoder and predictor architectures
    unchanged.

    \noindent\textbf{Prediction objective.}
    The action-conditioned latent predictor operates on the fused representation,
    \begin{equation}
        \hat{\mathbf z}_{t+1}
        =
        g_\psi\!\left(
            \mathbf z_{t-h+1:t},
            \mathbf a_{t-h+1:t}
        \right),
    \end{equation}
    and is trained with the mean-squared next-latent prediction loss
    \begin{equation}
        \mathcal L_{\mathrm{pred}}
        =
        \operatorname{MSE}
        \left(
            \hat{\mathbf z}_{t+1},
            \mathbf z_{t+1}
        \right).
        \label{eq:app_prediction_loss}
    \end{equation}

    \noindent\textbf{SIGReg and training objectives.}
    We use the original SIGReg implementation of LeWM without modification.
    SIGReg evaluates the Epps--Pulley normality criterion on random
    one-dimensional projections of the latent representations, encouraging each
    projected marginal toward a standard Gaussian. We use $J=1024$ random
    projection directions, following LeWM, and evaluate the regularizer
    independently at each temporal position across the training batch.

    Raw LeWM applies SIGReg directly to the latent representations:
    \begin{equation}
        \mathcal L_{\mathrm{Raw}}
        =
        \mathcal L_{\mathrm{pred}}
        +
        \lambda\,
        \mathcal L_{\mathrm{SIGReg}}(\mathcal Z).
        \label{eq:app_raw_objective}
    \end{equation}
    For TC-LeWM, temporal centering is applied to the same latent sequence,
    \begin{equation}
        \bar{\mathbf z}_t
        =
        \frac{1}{|\mathcal W_t|}
        \sum_{s\in\mathcal W_t}
        \mathbf z_s,
        \qquad
        \mathbf r_t
        =
        \mathbf z_t-\bar{\mathbf z}_t,
        \label{eq:app_training_residual}
    \end{equation}
    and the objective becomes
    \begin{equation}
        \mathcal L_{\mathrm{TC}}
        =
        \mathcal L_{\mathrm{pred}}
        +
        \lambda\,
        \mathcal L_{\mathrm{SIGReg}}(\mathcal R).
        \label{eq:app_tc_objective}
    \end{equation}
    Thus, Raw LeWM and TC-LeWM use the same architecture, prediction objective,
    and multi-view treatment, differing only in the SIGReg target:
    $\mathcal Z$ versus $\mathcal R$.

    \noindent\textbf{Behavior-cloning head.}
    After world model training, the encoder is frozen and used with a
    task-conditioned flow-matching behavior-cloning policy.
    For each camera view, we retain the CLS token together with a
    $4\times4$ spatially pooled grid of patch tokens from the frozen ViT encoder.
    The tokens from both views are concatenated into a visual token sequence
    $\mathbf V_t$ and provided to a DiT-style action head~\citep{peebles2023scalable}.

    The policy predicts an action chunk
    \begin{equation}
        \mathbf A_t
        =
        [
            \mathbf a_t,
            \ldots,
            \mathbf a_{t+H-1}
        ]
        \in\mathbb R^{H\times 7}.
    \end{equation}
    Action tokens are processed by self-attention and cross-attend to
    $\mathbf V_t$, while the flow time and task identity condition the
    transformer blocks. We train the policy with the flow-matching objective
    \begin{equation}
        \mathbf X_\gamma
        =
        (1-\gamma)\boldsymbol{\xi}
        +
        \gamma\mathbf A_t,
        \qquad
        \mathcal L_{\mathrm{BC}}
        =
        \left\|
            v_\phi(
                \mathbf X_\gamma,
                \gamma,
                \mathbf V_t,
                i
            )
            -
            (\mathbf A_t-\boldsymbol{\xi})
        \right\|_2^2,
        \label{eq:app_bc}
    \end{equation}
    where
    $\boldsymbol{\xi}\sim\mathcal N(\mathbf 0,\mathbf I)$,
    $\gamma\sim\mathcal U(0,1)$, and $i$ denotes the task identity.
    Actions are normalized using the training-set mean and standard deviation.
    During evaluation, action chunks are generated from Gaussian noise using
    10 Euler integration steps and mapped back to the original action scale.
    Raw LeWM and TC-LeWM use the same policy architecture and training protocol.

    \noindent\textbf{Training hyperparameters.}
    Table~\ref{tab:hyperparameters} summarizes the main training hyperparameters
    used in our experiments.

    \begin{table}[t]
        \centering
        \caption{Main training hyperparameters used in our experiments.}
        \label{tab:hyperparameters}
        \small
        \setlength{\tabcolsep}{5pt}
        \begin{tabular}{lll}
            \toprule
            Stage & Hyperparameter & Value \\
            \midrule
            \multirow{9}{*}{World model}
            & Architecture & ViT-Tiny \\
            & Frame skip   & 4 \\
            & Predictor context & $h=3$ \\
            & SIGReg projections & $J=1024$ \\
            & SIGReg weight $\lambda$
            & $0.09$ \\
            & Centering window & $W=4$ (TC-LeWM) \\
            & Optimizer & AdamW, lr $5\times10^{-5}$ \\
            & Batch size & 128 \\
            & Training steps & 10k (30k for 40-task) \\
            \midrule
            \multirow{6}{*}{BC policy}
            & Architecture & DiT \\
            & Conditioning & 34 visual tokens via cross-attention + task embedding \\
            & Action horizon & $H=8$ \\
            & Optimizer & AdamW, lr $2\times10^{-4}$ \\
            & Batch size & 256 \\
            & Training steps & 40k (4k for single-task) \\
            & Inference & 10 Euler steps \\
            \bottomrule
        \end{tabular}
    \end{table}

    \subsection{Monte Carlo Analysis Details}
    \label{app:mc_details}

    \noindent\textbf{Projected mixture model.}
    Following the projected temporal decomposition in
    Sec.~\ref{sec:mc_analysis},
    \begin{equation}
        x
        =
        \mathbf{w}^{\top}\mathbf{z}_t
        =
        \underbrace{\mathbf{w}^{\top}\bar{\mathbf{z}}_t}_{c}
        +
        \underbrace{\mathbf{w}^{\top}\mathbf{r}_t}_{\varepsilon},
        \label{eq:app_projected_decomposition}
    \end{equation}
    we construct a finite-window Monte Carlo surrogate in which each temporal
    window has a shared component center and independent within-window
    variation. Specifically, before temporal centering, projected samples are
    generated as
    \begin{equation}
        x_t
        =
        c_0 + \eta_t,
        \qquad
        c_0 \sim
        \frac{1}{K}\sum_{k=1}^{K}
        \mathcal{N}(\mu_k,\tau^2),
        \qquad
        \eta_t \sim \mathcal{N}(0,\sigma_r^2),
        \qquad
        \operatorname{Cov}(c_0,\eta_t)=0.
        \label{eq:app_mc_model}
    \end{equation}
    Here, $c_0$ is shared by all samples within a temporal window and
    $\eta_t$ represents independent within-window variation.
    Temporal centering gives
    \begin{equation}
        c
        =
        \bar{x}
        =
        c_0+\bar{\eta},
        \qquad
        \varepsilon_t
        =
        x_t-\bar{x}
        =
        \eta_t-\bar{\eta}.
        \label{eq:app_mc_centering}
    \end{equation}

    In the Monte Carlo analysis, we set $\tau=0$ so that the variance of the
    shared component is determined entirely by the spread of the component
    centers. We denote the two controlled variance parameters by
    \begin{equation}
        v_c
        \triangleq
        \mathrm{var}(c_0),
        \qquad
        v_\varepsilon
        \triangleq
        \mathrm{var}(\eta_t)
        =
        \sigma_r^2.
        \label{eq:app_mc_variances}
    \end{equation}
    For compactness, these controlled parameters are denoted by
    $\mathrm{var}(c)$ and $\mathrm{var}(\varepsilon)$ in the main-text
    Monte Carlo analysis and Fig.~\ref{fig:mechanism}(a--b).
    Their relation to the finite-window centered quantities is given below.

    \noindent\textbf{Randomized center construction.}
    For each Monte Carlo layout $m$, we sample $K$ raw component centers
    independently from a uniform distribution,
    \begin{equation}
        \widetilde{\mu}_k^{(m)}
        \overset{\mathrm{i.i.d.}}{\sim}
        \mathcal U(-1,1),
        \qquad
        k=1,\ldots,K.
    \end{equation}
    We then center and normalize the sampled centers to obtain
    $\{\widehat{\mu}_k^{(m)}\}_{k=1}^{K}$ with zero mean and unit variance.
    For a target persistent variance $v_c$, we rescale them as
    \begin{equation}
        \mu_k^{(m)}(v_c)
        =
        \sqrt{v_c}\,
        \widehat{\mu}_k^{(m)}.
        \label{eq:app_center_rescaling}
    \end{equation}
    Thus, each layout satisfies
    $\frac{1}{K}\sum_k \left(\mu_k^{(m)}\right)^2=v_c$.
    The normalized centers determine the relative component geometry, while
    $v_c$ directly controls its overall variance.

    \noindent\textbf{Monte Carlo protocol.}
    We use $K=10$ components, matching the number of tasks in each LIBERO
    suite. Each batch contains $N=128$ projected samples grouped into temporal
    windows of length $W=4$. Each window is assigned to one component, and its
    samples are generated as
    \begin{equation}
        x_t
        =
        \mu_k
        +
        \sqrt{v_\varepsilon}\,\xi_t,
        \qquad
        \xi_t\sim\mathcal N(0,1),
        \label{eq:app_mc_generation}
    \end{equation}
    where $v_\varepsilon$ controls the pre-centering within-window variation.
    We average over $M=400$ randomized center layouts and $R=20$ finite-batch
    realizations per layout. The same component assignments and Gaussian samples
    are reused across the variance grid for paired comparisons.

    For Fig.~\ref{fig:mechanism}(a), we sweep
    $v_c\in[0,4.2]$ with step $0.1$ and
    $v_\varepsilon\in[0.05,2.5]$ with step $0.05$.
    We instantiate the Raw LeWM objective in the projected Monte Carlo model as
    \begin{equation}
        \mathcal L_{\mathrm{Raw}}^{\mathrm{MC}}
        =
        v_\varepsilon
        +
        \lambda\,
        \mathcal L_{\mathrm{EP}}(x),
        \qquad
        \lambda=0.09.
        \label{eq:app_mc_objective}
    \end{equation}
    Here, $v_\varepsilon$ serves as a surrogate for
    $\mathcal L_{\mathrm{pred}}$, while
    $\mathcal L_{\mathrm{EP}}(x)$ corresponds to the SIGReg term evaluated on
    the one-dimensional projected marginal $x$.
    The resulting objective landscape is shown in
    Fig.~\ref{fig:mechanism}(a).
    This provides a scale-consistent surrogate for the original objective:
    the prediction term is represented by the scale of within-window variation
    along a projected direction, while the EP term is evaluated on the
    corresponding one-dimensional projected marginal.

    \noindent\textbf{EP loss under temporal centering.}
    For Fig.~\ref{fig:mechanism}(b), we fix
    $v_\varepsilon=1.30$ and vary only the persistent variance $v_c$.
    For $W=4$, temporal centering converts this to a centered-residual variance
    of
    \begin{equation}
        \mathrm{var}(\varepsilon)
        =
        \frac{3}{4}\times1.30
        =
        0.975,
    \end{equation}
    which is approximately one.
    For each setting, we evaluate the change in EP loss relative to
    $v_c=0$ on both the projected latent $x$ and its temporally centered
    residual.
    Because the component center is shared by all samples within a temporal
    window, it is removed exactly by temporal centering.
    Consequently, the residual EP loss is invariant to $v_c$ under this
    construction, whereas the EP loss on $x$ changes with persistent variance.

    \noindent\textbf{Finite-window variance conversion.}
    The Monte Carlo parameters $(v_c,v_\varepsilon)$ control the variance of
    the window-shared component $c_0$ and the pre-centering within-window
    variation $\eta_t$, respectively. The temporal decomposition in the main
    text instead measures the finite-window projected mean
    $c=\bar{x}$ and centered residual
    $\varepsilon_t=x_t-\bar{x}$.
    For independent $\eta_t$ within a window of length $W$,
    \begin{equation}
        \mathrm{var}(c)
        =
        \mathrm{var}(\bar{x})
        =
        v_c+\frac{1}{W}v_\varepsilon,
        \qquad
        \mathrm{var}(\varepsilon)
        =
        \frac{W-1}{W}v_\varepsilon.
        \label{eq:app_finite_window}
    \end{equation}
    Therefore, for the $W=4$ windows used in our experiments,
    \begin{equation}
        v_\varepsilon
        =
        \frac{4}{3}\mathrm{var}(\varepsilon),
        \qquad
        v_c
        =
        \mathrm{var}(c)
        -
        \frac{1}{3}\mathrm{var}(\varepsilon).
        \label{eq:app_variance_conversion}
    \end{equation}
    At the representation level,
    $\mathrm{var}(c)$ and $\mathrm{var}(\varepsilon)$ correspond to the
    per-dimension projected counterparts of
    $\mathrm{var}(\bar{\mathbf z})$ and
    $\mathrm{var}(\mathbf r)$, respectively.
    We use~\eqref{eq:app_variance_conversion} when mapping trained-system
    measurements onto the Monte Carlo coordinates in
    Fig.~\ref{fig:mechanism}(a).
    Figure~\ref{fig:mechanism}(c) instead reports the directly measured
    $\mathrm{var}(\bar{\mathbf z})$ and
    $\mathrm{var}(\mathbf r)$.

    \subsection{Representation Analysis Details}
    \label{app:representation_details}

    \begin{figure*}[t]
        \centering
        \includegraphics[width=\textwidth]{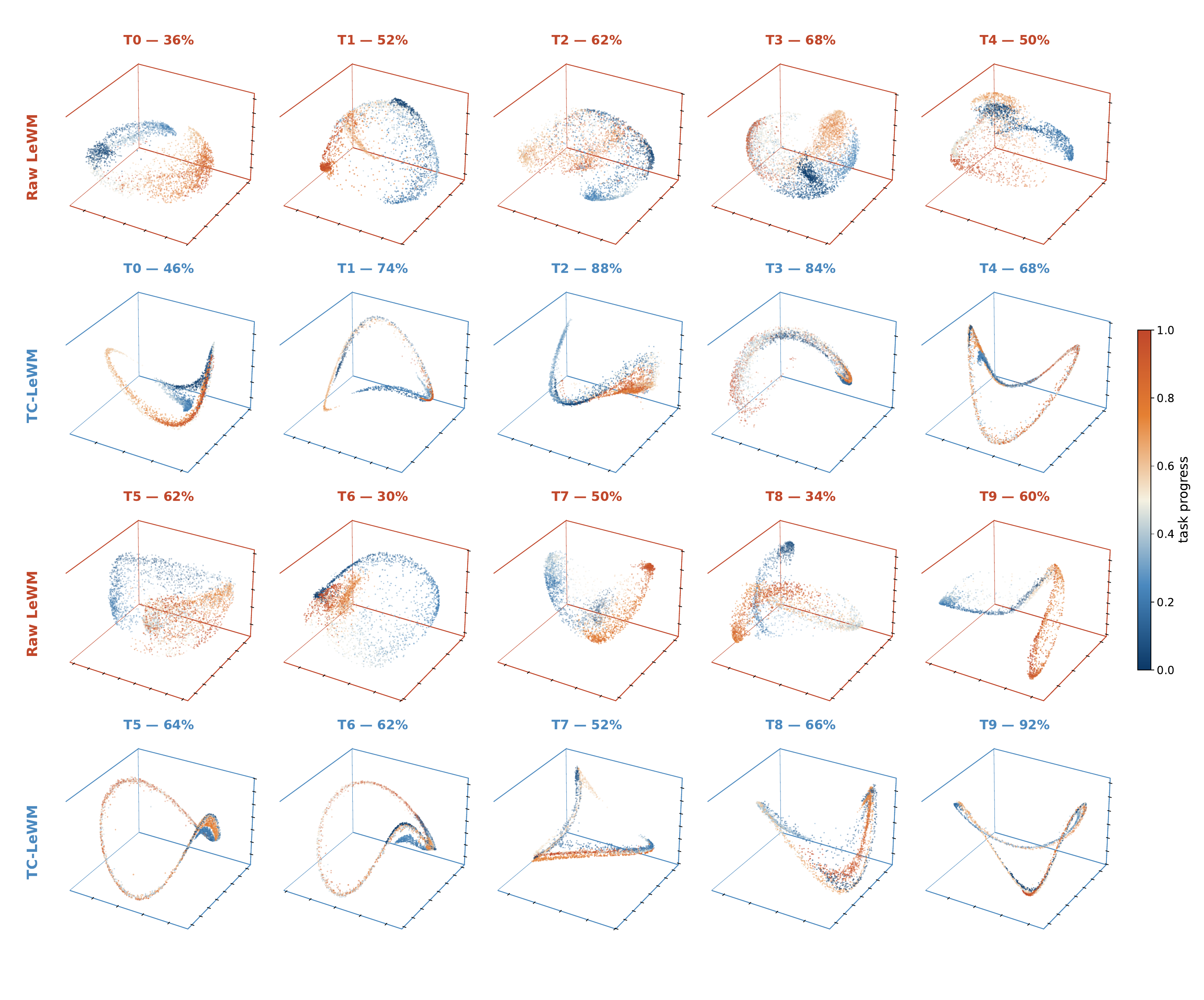}
        \caption{
            \textbf{Per-task 3D PCA visualization of latent representations} on LIBERO-Long
            under suite-wise 10-task training.
            Points are colored by normalized within-trajectory progress.
            Across many tasks, TC-LeWM exhibits smoother and more clearly
            progress-aligned structures, whereas Raw LeWM more often produces
            diffuse geometry.
            The percentages in the titles indicate the corresponding per-task
            closed-loop success rates for the displayed training seed, averaged over
            three evaluation seeds.
        }
        \label{fig:app_task_pca}
    \end{figure*}

    \noindent\textbf{Empirical variance allocation.}
    For Fig.~\ref{fig:analysis}(a), we apply the same empirical variance
    calculation to the projector outputs at each of the three training scales.
    Consecutive cached representations are grouped into windows of four frames,
    matching the temporal windows used during world model training.
    For each window, we compute the persistent component
    $\bar{\mathbf z}_t$ and centered residual $\mathbf r_t$ using the
    decomposition defined in the main text.

    We compute the empirical variance independently for each latent dimension
    and then average across dimensions to obtain
    $\mathrm{var}(\bar{\mathbf z})$ and $\mathrm{var}(\mathbf r)$.
    The resulting quantities are therefore reported on the same per-dimension
    variance scale as the unit-variance target of SIGReg.
    The same calculation is used across all training scales, and the displayed
    values are averaged over three independently trained seeds.

    \noindent\textbf{Latent content attribution.}
    For Fig.~\ref{fig:analysis}(b), we quantify how latent variation is
    associated with task identity, within-episode progress, and trajectory
    identity.
    For each evaluation cell, we first perform PCA on the frozen latent
    representations and retain the top eight principal components.
    We then apply sequential ANOVA independently to each principal component,
    using the fixed order task identity $\rightarrow$ within-episode progress
    $\rightarrow$ trajectory identity.
    Within-episode progress is discretized into 10 ordered progress stages.
    Under this ordering, trajectory identity captures only variation beyond that
    already explained by task identity and within-episode progress, avoiding
    double counting of progress-related variation.

    The attribution for each principal component is weighted by its
    explained-variance ratio before aggregation across the top eight components.
    Figure~\ref{fig:analysis}(b) reports the TC-LeWM minus Raw LeWM difference
    in the resulting variance share, such that positive values indicate that
    TC-LeWM allocates relatively more latent variation to the corresponding
    factor.
    The $4.5\times$, $2.5\times$, and $6.9\times$ annotations report the
    corresponding Raw-to-TC ratios of the trajectory-identity share.

    \noindent\textbf{Linear decodability of robot state and dynamics.}
    For Fig.~\ref{fig:analysis}(c), we evaluate how linearly decodable robot
    state and dynamics are from the frozen representations.
    The analysis is performed separately for each task, with $20\%$ of complete
    episodes held out for evaluation.

    We use ridge regression with regularization coefficient $\lambda=0.1$.
    Both latent inputs and regression targets are z-score normalized using
    statistics from the fitting split.
    State is decoded directly from $\mathbf z_t$.
    For velocity, we decode the first-order latent difference
    $\mathbf z_{t+1}-\mathbf z_t$ into the corresponding first-order robot-state
    difference, while acceleration analogously uses second-order temporal
    differences.

    Decodability is measured using Pearson correlation on the held-out episodes.
    Pearson correlation is computed independently for each output dimension on 
    the held-out episodes and then averaged across dimensions.
    The reported points are averaged over the four LIBERO suites and three
    independently trained seeds, with error bars indicating the standard
    deviation across training seeds.

    \noindent\textbf{Stage-wise closed-loop analysis.}
    For Fig.~\ref{fig:analysis}(d), we decompose each rollout into grasping and
    placement stages using simulator object states. For each task, we identify
    the movable object(s) of interest from the task definition and track
    their 3D positions throughout evaluation. A rollout is considered to have
    passed the grasping stage if any target object is lifted by at least
    $3\,\mathrm{cm}$ relative to its settled initial height. The placement-stage
    success rate is then defined conditionally as the fraction of grasp-successful
    rollouts that eventually complete the task. 

    We apply this decomposition only to tasks whose successful executions involve
    lifting a target object. In LIBERO-Goal, three tasks do not satisfy this
    criterion and are therefore excluded from the stage-wise analysis, leaving
    seven grasp-and-place tasks. 

    \subsection{Additional Latent-Space Analysis}
    \label{app:additional_latent_analysis}

    \noindent\textbf{Per-task PCA visualization.}
    For the frozen LIBERO-Long representations learned under suite-wise
    10-task training, we perform PCA separately for each task and visualize
    the first three principal components.
    Figure~\ref{fig:app_task_pca} compares Raw LeWM and TC-LeWM on one training seed, with points
    colored by normalized within-trajectory progress.
    Across many tasks, TC-LeWM exhibits connected and more clearly
    progress-aligned structures, whereas Raw LeWM more often produces
    diffuse shapes.
    In the per-task comparison between Raw LeWM and TC-LeWM,
    improvements in closed-loop success are often accompanied by more organized
    and manifold-like latent trajectories.

    \noindent\textbf{Gaussianity of the full latent.}
    After world model training, we freeze the encoder and evaluate the EP loss
    of the resulting full latent representation $\mathbf z$ under suite-wise
    10-task training.
    As shown in Table~\ref{tab:app_ep_z}, Raw LeWM consistently exhibits a much
    lower EP loss, as expected because SIGReg directly regularizes the full latent
    toward the Gaussian target during training.
    In contrast, TC-LeWM applies SIGReg only to the centered residual and removes
    the Gaussian constraint from the full latent $\mathbf z$.
    Consequently, the frozen TC-LeWM latent deviates more strongly from the Gaussian target.
    This higher EP loss is therefore an intended consequence of the method:
    releasing the full latent from the Gaussian target allows structured and
    potentially multimodal organization to emerge in $\mathbf z$, while the
    centered residual remains the target of SIGReg.

\begin{figure*}[t]
        \centering
        \includegraphics[width=\textwidth]{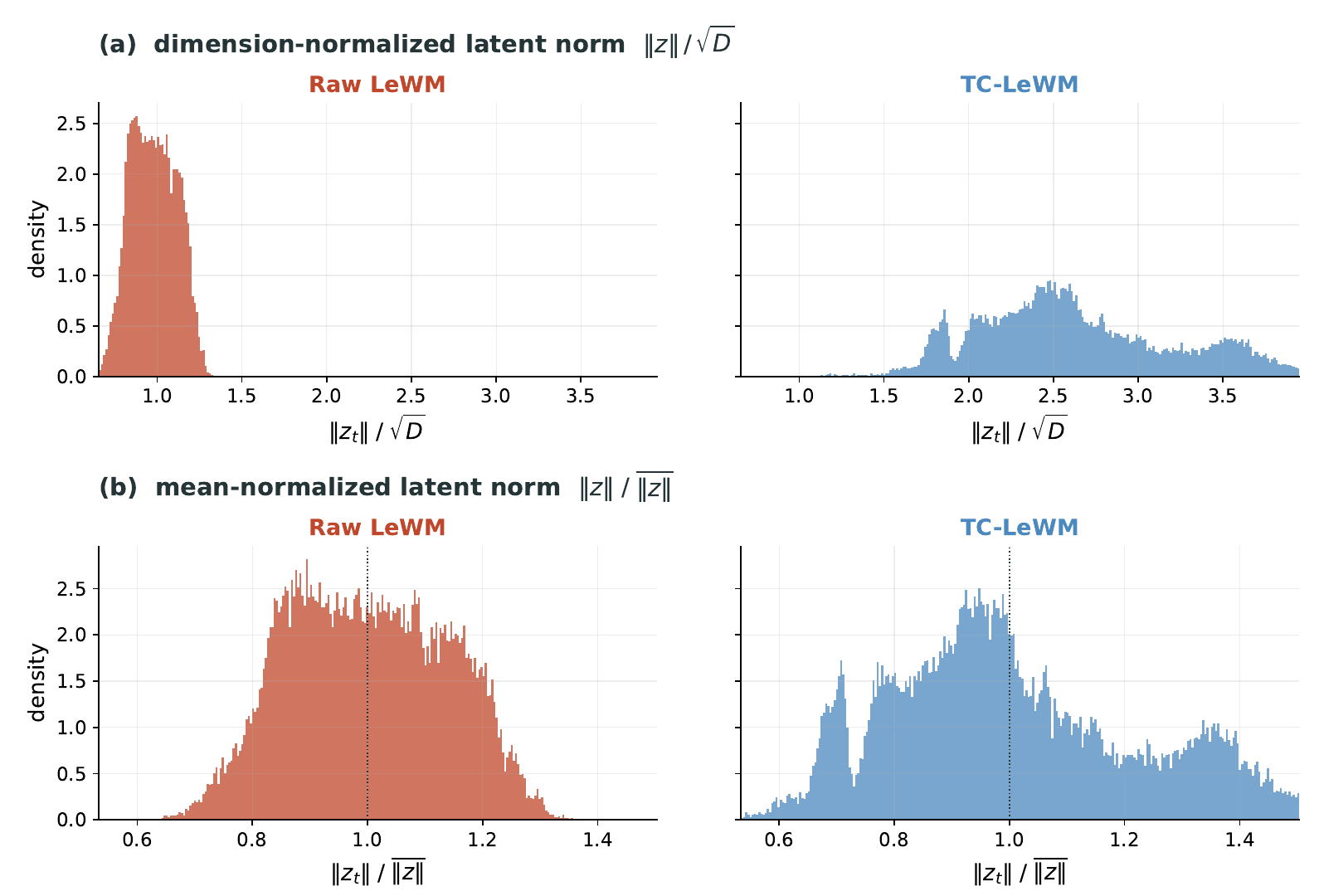}
        \caption{
            Latent norm distributions on LIBERO-Long under 10-task training.
            (a) Dimension-normalized latent norm
            $\|\mathbf z_t\|/\sqrt{D}$, which preserves the relative latent scale
            while using the typical norm of a high-dimensional standard Gaussian
            as a reference near one.
            Raw LeWM remains concentrated near this Gaussian radial scale, whereas
            TC-LeWM spans a substantially broader range with multiple radial modes.
            (b) Mean-normalized latent norm
            $\|\mathbf z_t\|/\overline{\|\mathbf z\|}$, which removes the difference
            in average scale between the two representations.
            The multimodal radial structure of TC-LeWM remains clearly visible after
            normalization, showing that the difference cannot be explained solely
            by global rescaling.
        }
        \label{fig:app_norm_hist}
    \end{figure*}

    \noindent\textbf{Latent radial structure.}
    To further characterize the non-Gaussian structure of the full latent,
    Figure~\ref{fig:app_norm_hist} compares the latent norm distributions of
    Raw LeWM and TC-LeWM on LIBERO-Long under 10-task training.
    We first report the dimension-normalized norm
    $\|\mathbf z_t\|/\sqrt{D}$, where $D$ is the latent dimension.
    This normalization preserves the relative scale difference between the two
    methods while providing a natural reference: for a high-dimensional
    standard isotropic Gaussian, the norm concentrates near $\sqrt{D}$, and
    hence $\|\mathbf z_t\|/\sqrt{D}$ is concentrated near one.
    Consistent with its low EP loss, Raw LeWM is tightly concentrated around
    this Gaussian radial scale, whereas TC-LeWM spans a substantially broader
    range and exhibits multiple radial modes.

    To separate this structural difference from a simple change in global scale,
    we additionally normalize each representation by its own mean norm,
    $\|\mathbf z_t\|/\overline{\|\mathbf z\|}$.
    After removing the difference in average scale, TC-LeWM still exhibits a
    clearly more multimodal radial distribution, while Raw LeWM remains largely
    unimodal.
    These results indicate that releasing the full latent from the Gaussian
    target changes not only its overall radial scale but also its distributional
    shape, allowing richer radial organization to emerge.

    \begin{table}[t]
        \centering
        \caption{
            EP loss evaluated on the full latent representation $\mathbf z$
            extracted from the frozen encoder after suite-wise 10-task training.
            Results are mean $\pm$ standard deviation over three training seeds
            and are reported in units of $10^{-2}$.
            Lower values indicate a distribution closer to the standard Gaussian target.
        }
        \label{tab:app_ep_z}
        \setlength{\tabcolsep}{4pt}
        \begin{tabular}{lccccc}
            \toprule
            Method & Object & Spatial & Goal & Long & Avg. \\
            \midrule
            Raw LeWM
            & $0.45 \pm 0.02$
            & $0.39 \pm 0.05$
            & $0.40 \pm 0.01$
            & $0.39 \pm 0.02$
            & $0.41$ \\
            TC-LeWM
            & $1.89 \pm 0.10$
            & $2.16 \pm 0.96$
            & $2.24 \pm 0.15$
            & $2.66 \pm 1.36$
            & $2.24$ \\
            \bottomrule
        \end{tabular}
    \end{table}

    \begin{table}[t]
        \centering
        \caption{
            Sensitivity of TC-LeWM to the temporal-centering window size $W$
            on LIBERO-Long under suite-wise 10-task training.
            Results are mean$\pm$standard deviation over three training seeds.
            The temporal context used by the predictor and the remaining training
            setup are kept fixed.
        }
        \label{tab:window_ablation}
        \small
        \setlength{\tabcolsep}{5pt}
        \begin{tabular}{lcccc}
            \toprule
            $W$ & 2 & 4 (default) & 8 & 16 \\
            \midrule
            Success (\%)
            & $62.0\pm3.4$
            & $63.6\pm6.8$
            & $\mathbf{65.8\pm3.5}$
            & $52.1\pm3.8$ \\
            \bottomrule
        \end{tabular}
    \end{table}

    \begin{table}[t]
        \centering
        \caption{
            Sensitivity of Raw LeWM to the SIGReg weight $\lambda$
            on LIBERO-Long under suite-wise 10-task training.
            Results are mean$\pm$standard deviation over three training seeds.
            The model architecture and the remaining training
            setup are kept fixed.
        }
        \label{tab:lambda_ablation}
        \small
        \setlength{\tabcolsep}{5pt}
        \begin{tabular}{lccccc}
            \toprule
            $\lambda$ & 0.005 & 0.01 & 0.03 & 0.06 & 0.09 \\
            \midrule
            Success (\%)
            & $43.4\pm1.5$
            & $\mathbf{50.3\pm6.8}$
            & $49.5\pm2.9$
            & $47.1\pm0.6$
            & $44.7\pm4.3$ \\
            \bottomrule
        \end{tabular}
    \end{table}

    \subsection{Ablation Studies}
    \label{app:ablations}

    \noindent\textbf{Sensitivity to temporal window size.}
    We evaluate the sensitivity of TC-LeWM to the temporal-centering window
    size $W$ on LIBERO-Long under suite-wise 10-task training.
    As shown in Table~\ref{tab:window_ablation}, performance is similar for
    $W=2$, $4$, and $8$, with average success rates of $62.0\%$, $63.6\%$,
    and $65.8\%$, respectively. This suggests that TC-LeWM is not particularly
    sensitive to the precise window size within a relatively short temporal
    range. Performance decreases to $52.1\%$ when the window is increased to
    $W=16$, indicating that overly long windows are less suitable for the local
    temporal decomposition.

    We use $W=4$ by default because it matches the four-frame training sequence
    used to organize world model training batches.
    We do not evaluate $W=1$, since temporal centering then gives
    $\mathbf r_t=\mathbf 0$ identically and therefore does not provide a
    meaningful residual regularization target.

    \noindent\textbf{Sensitivity to SIGReg strength in Raw LeWM.}
    We further examine whether the downstream limitation of Raw LeWM can be
    addressed simply by reducing the SIGReg strength.
    As shown in Table~\ref{tab:lambda_ablation}, moderately decreasing $\lambda$
    from the default value of $0.09$ improves downstream success, with the best
    performance obtained at $\lambda=0.01$.
    However, further reducing the regularization strength to $\lambda=0.005$
    degrades performance, indicating that the improvement does not continue as
    the regularization pressure is reduced.
    Moreover, even the best Raw LeWM result remains substantially below TC-LeWM.
    These results suggest that reducing the regularization pressure alone cannot
    fully resolve the downstream representation limitation of Raw LeWM.

\end{document}